\documentclass[11pt,letterpaper]{article}

\usepackage{graphicx}%
\usepackage{multirow}%
\usepackage{amsmath,amssymb,amsfonts}%

\usepackage{amsthm}%
\usepackage{mathrsfs}%
\usepackage[title]{appendix}%
\usepackage{xcolor}%
\usepackage{textcomp}%
\usepackage{manyfoot}%
\usepackage{booktabs}%
\usepackage{algorithm}%
\usepackage{algorithmicx}%
\usepackage{algpseudocode}%
\usepackage{listings}%
\usepackage[skip=10pt plus1pt, indent=6pt]{parskip}

\usepackage{diagbox}
\usepackage{bm}
\usepackage{makecell}
\usepackage[usestackEOL]{stackengine}
\usepackage[colorlinks=true, allcolors=blue]{hyperref}

\AtBeginEnvironment{appendices}{\crefalias{section}{appendix}}

\usepackage{arydshln}
\usepackage{stackengine}
\usepackage[margin = 1 in]{geometry}
\usepackage{mathtools}
\usepackage{bbm, dsfont}
\usepackage[font={footnotesize}, labelfont={bf}, labelsep=space, justification=justified, skip=0pt]{caption}
\usepackage[hang,flushmargin]{footmisc} 
\usepackage{subcaption}
\usepackage{placeins}

\usepackage{lineno}

\usepackage{titling}
\thanksheadextra{}{}
\thanksheadextra{}{} 
\usepackage{authblk}

\usepackage{cleveref} 
\usepackage{cite} 

\newtheorem{theorem}{Theorem}[section]

\theoremstyle{definition}
\newtheorem{definition}{Definition}[section]
\newcommand\Rey{\mbox{\textit{Re}}}  

\newcommand{\cmt}[1]{} 
\newcommand{\ie}{i.e., }
\newcommand{\eg}{e.g., }

\newcommand{\name}{Corrective Residuals}
\newcommand{\short}{CoRes}
\newcommand{\shortspace}{CoRes }
\newcommand{\GPor}{GP$_\text{OR}$}
\newcommand{\GPorSpace}{GP$_\text{OR}$ }
\newcommand{\PINNdw}{PINN$_\text{DW}$}
\newcommand{\PINNhc}{PINN$_\text{HC}$}

\newcommand{\lt}{\mathcal{L}(\boldsymbol{\theta})}

\newcommand{\zb}{\boldsymbol{z}}

\newcommand{\inputb}{\mathbf{x}}
\newcommand{\Inputb}{\mathbf{X}}

\newcommand{\rb}{\boldsymbol{r}}
\newcommand{\wb}{\boldsymbol{w}}
\newcommand{\Cb}{\boldsymbol{C}}

\newcommand{\lambdab}{\boldsymbol{\lambda}}

\newcommand{\inputspace}{\mathcal{X}}
\newcommand{\hsspace}{\mathcal{H}}
\newcommand{\rspace}{\mathbb{R}}
\newcommand{\expectation}{\mathbb{E}}

\newcommand{\diag}{\text{diag}}

\newcommand{\hb}{\boldsymbol{h}}

\newcommand{\thetab}{\boldsymbol{\theta}}
\newcommand{\phib}{\boldsymbol{\phi}}

\newcommand{\omegab}{\boldsymbol{\omega}}
\newcommand\brackets[1]{\mathopen{}\left[#1\right]\mathclose{}}
\newcommand\parens[1]{\mathopen{}\left(#1\right)\mathclose{}}
\newcommand\braces[1]{\mathopen{}\left\{#1\right\}\mathclose{}}

\newcommand\bars[1]{\left|#1\right|}

\newcommand{\outputu}{\text{u}}
\newcommand{\outputub}{\mathbf{u}}

\newcommand{\ut}{u_t}
\newcommand{\ux}{u_x}
\newcommand{\uy}{u_y}
\newcommand{\uxx}{u_{xx}}
\newcommand{\uyy}{u_{yy}}

\newcommand{\vx}{v_{x}}
\newcommand{\vy}{v_{y}}
\newcommand{\vxx}{v_{xx}}
\newcommand{\vyy}{v_{yy}}

\newcommand{\px}{p_{x}}
\newcommand{\py}{p_{y}}

\title{A Gaussian Process Framework for Solving Forward and Inverse Problems Involving Nonlinear Partial Differential Equations}
\date{\vspace{-5ex}}
\author[1]{Carlos Mora}
\author[1]{Amin Yousefpour}
\author[1]{Shirin Hosseinmardi}
\author[1]{Ramin Bostanabad\thanks{Corresponding Author: Raminb@uci.edu}}
\affil[1]{Department of Mechanical and Aerospace Engineering, University of California, Irvine}

\begin{document}
    \pagenumbering{arabic}
    \sloppy    
    \maketitle
    \noindent \textbf{Abstract}\\Physics-informed machine learning (PIML) has emerged as a promising alternative to conventional numerical methods for solving partial differential equations (PDEs). 
PIML models are increasingly built via deep neural networks (NNs) whose architecture and training process are designed such that the network satisfies the PDE system. 
While such PIML models have been substantially advanced over the past few years, their performance is still very sensitive to the network's architecture, loss function, and optimization settings. 
Motivated by this limitation, we introduce kernel-weighted Corrective Residuals (\short) to integrate the strengths of kernel methods and deep NNs for solving nonlinear PDE systems. To achieve this integration, we design a framework based on Gaussian processes (GPs) whose mean functions are parameterized via deep NNs. The resulting PIML model, abbreviated as NN-\short, can solve PDE systems without any labeled data inside the domain and is particularly attractive because it $(1)$ naturally satisfies the boundary and initial conditions of a PDE system in arbitrary domains, and $(2)$ can leverage any differentiable function approximator, \eg deep NN architectures, in its mean function. 
To ensure computational efficiency and robustness, we devise a modular approach for NN-\short\ to separately estimate the parameters of the kernel and the deep NN. Our studies indicate that NN-\shortspace consistently outperforms competing methods and considerably decreases the sensitivity of NNs to factors such as random initialization, architecture type, and choice of optimizer. We believe our findings have the potential to spark a renewed interest in leveraging kernel methods for solving PDEs\footnote{GitHub repository: \href{https://github.com/Bostanabad-Research-Group/GP-for-pde-solving}{https://github.com/Bostanabad-Research-Group/GP-for-pde-solving}}. 

    \section{Introduction} \label{sec: intro}

Partial differential equations (PDEs) elegantly explain the behavior of many engineered and natural systems such as power grids \cite{RN995, schaeffer2013sparse}, advanced materials \cite{RN1024}, tectonic cracks \cite{RN829}, weather and climate \cite{RN936, RN1311}, and biological agents \cite{RN995, santolini2018predicting}. 
Since the solutions to most PDEs cannot be derived analytically, numerical approaches such as the finite element method are frequently used to solve them. Recently, a new class of methods known as physics-informed machine learning (PIML) has been developed and successfully used in studying fundamental phenomena such as turbulence \cite{lucor2022simple}, diffusion \cite{fang2023physics}, shock waves \cite{jagtap2022physics}, interatomic bonds \cite{pun2019physically}, and cell signaling \cite{lotfollahi2023biologically}. 
While PIML models have fueled a renaissance in modeling complex systems, their performance heavily depends on optimizing the model's training mechanism (e.g., choice of optimizer), loss function, and architecture \cite{RN1958}.
To reduce the time and energy footprint of developing PIML models while improving their accuracy, we re-envision solving PDEs via machine learning (ML) and introduce a framework based on Gaussian processes (GPs) to simultaneously use the strengths of deep neural networks (NNs) and kernel methods. Specifically, our PIML model augments NNs with what we call kernel-weighted Corrective Residuals (\short) to improve NNs' accuracy, robustness, and efficiency in solving PDEs.

\subsection{Background on Physics-informed Machine Learning} \label{sec: piml}

Remarkable successes have been achieved via ML in many areas such as protein modeling \cite{kozuch2018combined, coin2003enhanced}, designing new materials \cite{curtarolo2013high, butler2018machine, hart2021machine, shi2019deep, RN666, liu2023special}, automated demographics monitoring \cite{gebru2017using}, and expert language models \cite{thirunavukarasu2023large}.
Availability of big data is a common feature across these applications which, in turn, enables building large ML models that can distill highly complex relations from the data. However, in the context of solving PDEs, there is a particular interest in building ML models whose training does not rely on any sample solutions inside the domain. 
Indeed, the key enabler in this application is PIML which systematically infuses our physical and mathematical knowledge into the structure and/or training mechanism of ML models. Compared to classical computational tools, PIML promises a unified platform for solving inverse problems \cite{RN1116}, obtaining meshfree solutions \cite{RN1637, RN1637, RN1929}, assimilating experiments with simulations \cite{von2022mean}, and uncertainty quantification \cite{RN1835}. 

We can broadly classify PIML models into two categories. The first group of methods relies on variants of neural networks (NNs) and can be traced back to \cite{djeridane2006neural, lagaris1998artificial}. 
Physics-informed neural networks (PINNs) \cite{raissi2019physics,RN1926} and their various extensions \cite{mcclenny2020self, jagtap2020adaptive, wang2021understanding, bu2021quadratic, gao2021phygeonet, jagtap2021extended} are perhaps the most widely adopted member of these classes of methods and their basic idea is to parameterize the PDE solution via a deep NN. As detailed in \Cref{sec: methods-description}, the parameters of this NN are optimized by minimizing a multi-component loss function which encourages the NN to satisfy the PDE as well as the initial and/or boundary conditions (ICs and BCs), see \Cref{fig: pinn-flowchart}. This minimization relies on automatic differentiation \cite{baydin2018automatic} and is known to be very sensitive to the optimizer, loss function formulation, and NN's architecture. To decrease this sensitivity, recent works have developed adaptive loss functions \cite{chen2018gradnorm, van2022optimally} and tailored architectures that improve gradient flows \cite{RN1530} or automatically satisfy BCs \cite{lagari2020systematic, lagaris1998artificial, dong2021method, RN1920, berg2018unified} (see \Cref{sec: methods-description} for more details). These advancements, however, fail to generalize to a diverse set of PDEs and substantially increase the cost and complexity of training \cite{RN1955}. 
Deep NNs have also been integrated with classical numerical solvers such as the finite element method (FEM) in various ways. For instance, HiDeNN-FEM \cite{zhang2021hierarchical} and its extensions \cite{lu2023convolution, zhang2022hidenn} utilize hierarchical neural architectures to learn more flexible and adaptive shape functions to achieve higher accuracy compared to the FEM and, in turn, benefit applications such as high-resolution topology optimization. These works focus on solving specific instances of a PDE system but NNs have also been used for operator learning \cite{lu2021learning,li2020fourier} where the idea is to approximate mappings between the inputs and outputs of a PDE system.

The second group of PIML models leverage kernel methods. The key idea behind kernel methods is to implicitly map the original input data into a higher dimensional space where it is easier to quantify similarities among the data points. Support vector machines (SVMs) are a popular kernel method that have been successfully applied to supervised learning, clustering, dimensionality reduction, and anomaly detection \cite{salcedo2014support}. GPs are also kernel methods and can be traced back to Poincaré’s course in probability theory \cite{owhadi2019statistical}. GPs have long been used in emulation and Bayesian optimization but their application in solving PDEs is relatively new and remains largely unexplored. The few existing works \cite{RN1873, RN1919, RN1890} exclusively employ zero-mean GPs which are completely characterized by their parametric kernel or covariance function. With this choice, solving the PDE amounts to designing the GP's kernel whose parameters are obtained via either maximum likelihood estimation (MLE) or a regularized MLE where the penalty term quantifies the GP's error in satisfying the PDE system.

In a recent novel work \cite{RN1886}, solving PDEs via a zero-mean GP is cast as an optimal recovery problem whose loss function is derived based on the PDE system and aims to estimate the solution at a finite number of interior nodes in the domain. Once these values are estimated, the PDE solution is approximated anywhere in the domain via kernel regression (see \Cref{sec: methods-description}). Hereafter, we denote this method as \GPorSpace and note that it has been recently extended to learn operators \cite{RN1881} and to handle large datasets using the concept of inducing points \cite{RN1890}. Additionally, while GPs with non-zero means have been employed for solving PDEs in a data-driven manner \cite{wang2023discovery}, they have not been explored in a purely physics-informed setting with a robust training mechanism.

\subsection{Outline of the Paper}
The rest of our paper is organized as follows. We introduce our approach in \Cref{sec: methodology-all} where we first provide a theoretical rationale for it in \Cref{sec: theoretical-rationale} and then in \Cref{sec: proposed-framework} introduce the modular and robust framework that we have developed for efficiently implementing it. We comment on the most prominent features of our approach in \Cref{sec: character-extension} where we also introduce its extensions for solving inverse problems or handling PDE systems with multiple outputs.
We rigorously study the performance of our approach in \Cref{sec: results} and conclude the paper with some final remarks and future research directions in \Cref{sec: conclusion}.

    \section{Neural Networks with Kernel-weighted} \name \label{sec: methodology-all}
Kernel methods, particularly GPs, have less extrapolation and scalability powers compared to deep NNs. They also struggle to approximate PDE solutions that have large gradients or involve coupled dependent variables such as the Navier-Stokes equations.
However, GPs locally generalize better than NNs \cite{RN1919} and are interpretable and easy to train (see \Cref{sec: gp-properties} for discussions and examples). 
Grounded on these properties, we introduce deep architectures with kernel-weighted \shortspace that integrate the attractive features of NNs and GPs for solving PDEs.


\subsection{Theoretical Rationale} \label{sec: theoretical-rationale}
GPs provide a tractable and robust framework for function approximation \cite{RN332, RN1886, RN1935} and their kernels are extensively studied to accommodate learning functions of varying degrees of complexity \cite{RN1927, RN783, RN940, RN1559, RN434, RN1912, RN1573}. 
However, we argue that the sole reliance on the kernel serves as a double-edged sword when solving PDEs via GPs. 
To demonstrate, we consider the task of emulating the function $\outputu(\inputb)$ given the $n$ samples $\Inputb=\braces{\inputb_1, \cdots, \inputb_n}$ with corresponding outputs $\outputub = \braces{\outputu_1, \cdots, \outputu_n}$ where $\inputb \in \inputspace \subset \rspace^{dx}$ with boundary $\partial \inputspace$ and $\outputu_i=\outputu(\inputb_i)\in \rspace$. 
If we endow $\outputu(\inputb)$ with a GP prior with the mean function $m\parens{\inputb; \thetab}$ and kernel $c\parens{\inputb, \inputb'; \phib}$, the conditional process is also a GP whose expected value at $\inputb^*$ is:
\begin{subequations}
    \begin{align}
        &\eta\parens{\inputb^*; \thetab, \phib} \coloneqq
        \expectation\brackets{\outputu^*|\outputub, \Inputb} = 
        m\parens{\inputb^*; \thetab} + \wb^T\rb, 
        \label{eq: gp-conditional-mean-1} \\
        & \wb \coloneqq w\parens{\inputb^*, \Inputb; \phib} = c^{-1}\parens{\Inputb, \Inputb; \phib}c\parens{\Inputb, \inputb^*; \phib} 
        \label{eq: weights}
        \\
        & \rb \coloneqq r\parens{\Inputb, \outputub; \thetab} = \outputub - m\parens{\Inputb; \thetab}.
        \label{eq: residuals}  
    \end{align}
    \label{eq: gp-conditional-mean}        
\end{subequations}
Here, $\thetab$ are the parameters of the mean function, $\phib$ are the so-called length-scale or roughness parameters of the kernel, $c\parens{\Inputb, \inputb^*; \phib}=\brackets{c\parens{\inputb_1, \inputb^*; \phib}, \cdots, c\parens{\inputb_n, \inputb^*; \phib}}^T$, $\rb$ denotes the \textit{residuals} of the mean function on the training data, $\wb$ are the kernel-induced weights, and $\Cb = c\parens{\Inputb, \Inputb; \phib}$ is the covariance matrix with $ij^{th}$ entry $c\parens{\inputb_i, \inputb_j; \phib}$. 
The covariance function can be a deep NN \cite{RN1901} or the \textit{simple} kernel:
\begin{equation} 
    \begin{split}
        c\parens{\inputb, \inputb'; \phib} = \exp\braces{-\parens{\inputb - \inputb'}^T\diag(\phib)\parens{\inputb - \inputb'}},
    \end{split}
    \label{eq: kernel-main}
\end{equation}
which is the simplified version of the Gaussian covariance function:
\begin{equation} 
    c\parens{\inputb, \inputb'; \sigma^2, \phib, \delta} = \sigma^2 \exp\braces{-\parens{\inputb - \inputb'}^T\diag(\phib)\parens{\inputb - \inputb'}} + \mathbbm{1}\{\inputb==\inputb'\}\delta,
    \label{eq: kernel-main2}
\end{equation}
where $\sigma^2$ is the process variance, $\mathbbm{1}\{\cdot\}$ returns $1/0$ if the enclosed statement is true/false, and $\delta$ is the so-called nugget parameter that is typically used to model noise or improve the numerical stability of the covariance matrix. For simplicity, hereafter we consider the Gaussian covariance function in \Cref{eq: kernel-main} in our descriptions. 

The optimum model parameters $\thetab$ and $\phib$ are generally unknown and hence estimated via MLE:
\begin{equation} 
    \begin{split}
        \brackets{\widehat\thetab, \widehat\phib} = 
        \underset{\thetab, \phib}{\operatorname{argmax}}\left|2 \pi \Cb\right|^{-\frac{1}{2}} 
        \exp \left\{\frac{-1}{2}\rb^T \Cb^{-1}\rb\right\},
    \end{split}
    \label{eq: mle}
\end{equation}
which can be an expensive and/or numerically unstable process if $\Cb$ is large or ill-conditioned (this can happen if the training dataset is large or has samples that are very close in the input space).

Since many kernels can approximate an arbitrary continuous function \cite{RN332}, zero-mean GPs are used in many regression problems as eliminating $m\parens{\inputb; \thetab}$ reduces the number of trainable parameters while increasing numerical stability. As discussed in \Cref{sec: gp-properties} and shown in \Cref{fig: gp-plots}, the latter improvement stems from the fact that an over-parameterized $m\parens{\inputb; \thetab}$, while needed for learning hidden complex relations, can easily interpolate $\outputub$ and, in turn, drive the residuals in \Cref{eq: residuals} to $\mathbf{0}$. Such residuals require $\phib\rightarrow\mathbf{0}$ which diminishes the contributions of the kernel in \Cref{eq: gp-conditional-mean-1} and renders $\Cb$ ill-conditioned.

Unlike regression, PDE systems cannot be accurately solved via zero-mean GPs without any in-domain samples since the posterior process in \Cref{eq: gp-conditional-mean} predicts zero for any point that is sufficiently far from the boundaries. This reversion to the mean behavior is due to the exponential decay of the correlations as the distance between two points increases, see \Cref{eq: kernel-main}. While deep kernels can delay this decay, they cannot prevent it. 

Following the above discussions, we make two important observations on the posterior distribution in \Cref{eq: gp-conditional-mean}: it heavily relies on $m\parens{\inputb; \thetab}$ in data scarce regions and it regresses $\outputub$ regardless of the values of $m\parens{\Inputb; \thetab}$. These observations suggest that a GP whose mean function is parameterized with a deep NN provides an attractive \textit{prior} for solving PDE systems since functions that are formulated as in \Cref{eq: gp-conditional-mean-1} can easily satisfy the BCs/IC and their smoothness can be controlled through the mean and covariance functions.
This approach, however, presents two major challenges.
First, the posterior distribution in this case should be obtained by conditioning the prior on BCs/IC while constraining it to satisfy the PDE in the domain. Since most practical PDEs are nonlinear, the posterior will not be Gaussian upon the constraining and hence there are no closed form formulas available for its likelihood (to train the model) or expected value (to easily predict with the model).
Second, jointly optimizing $\phib$ and $\thetab$ is a computationally expensive and unstable process due to the repeated need for constructing and inverting $\Cb$.

\subsection{Proposed Framework} \label{sec: proposed-framework}
\begin{figure}[!b]
    \centering
    \includegraphics[width=1.0\linewidth]{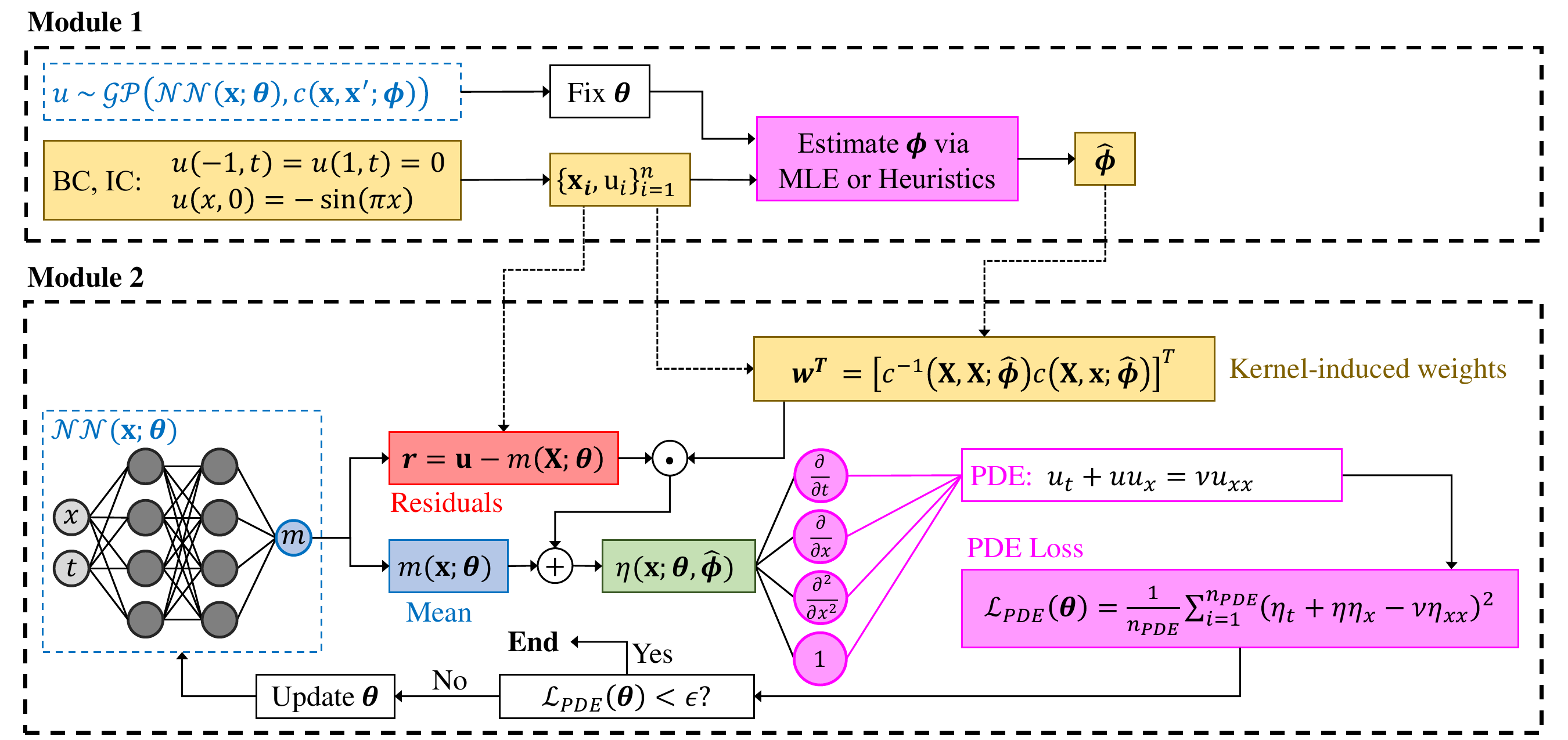}
    \caption{\textbf{Flowchart of the proposed framework for solving the 1D Burgers' equation:} We endow the solution $u(\inputb)$ with a GP prior whose mean and covariance functions are parameterized via a deep NN and the Gaussian kernel in \Cref{eq: kernel-main}, respectively. In module 1, we fix $\thetab$ to some random values and estimate the kernel parameters via MLE or heuristics such that the posterior GP conditioned on the BC/IC data faithfully reproduces $\textbf{u}$. Then, in Module 2, we estimate $\thetab$ by minimizing a loss function which only depends on the PDE since BC/IC are automatically satisfied.}
    \label{fig: flowchart}
\end{figure}

We address the above challenges via modularization and formulating the training process based on maximum a posteriori (MAP) instead of MLE. As illustrated in \Cref{fig: flowchart}, our framework consists of two sequential modules that aim to solve PDE systems with deep NNs that substantially benefit from kernel-weighted \short. These modules seamlessly integrate the best of two worlds: $(1)$ the local generalization power of kernels close to the domain boundaries where IC/BCs are specified, and $(2)$ the substantial capacity of deep NNs in learning multiple levels of distributed representations in the interior regions where there are no labeled training data.

In the first module, we endow the PDE solution with a GP prior whose mean and covariance functions are a deep NN and the Gaussian kernel in \Cref{eq: kernel-main}, respectively (note that the assumption on having a unit variance does not affect our method as the variance cancels out in \Cref{eq: weights} due to the matrix inversion). Conditioned on the data (i.e., $\outputub$ which is obtained by sampling from the BCs/IC), the posterior distribution of the solution is again a GP and follows \Cref{eq: gp-conditional-mean-1} where $\rb$ and $\wb$ denote the residuals and kernel-induced weights, respectively. Importantly, in this module we fix $\thetab$ to some random values and choose $\widehat\phib$ such that the GP can faithfully reproduce $\outputub$. As demonstrated in \Cref{sec: gp-properties} and \Cref{sec: sensitivity-analyses}, our results are not affected by the fact that we do not leverage MLE for parameter estimation in module one. Hence, we manually select $\widehat\phib$, fix the process variance to unity, and choose the nugget parameter to ensure the covariance matrix is not ill-conditioned (see \Cref{sec: gp-properties} for more details). 

In the second module, we obtain the final model by conditioning the GP on the (nonlinear) constraints that require the predictions at arbitrary points in the domain to satisfy the PDE system. We achieve this conditioning by fixing $\widehat\phib$ from module 1 and optimizing $\thetab$ to ensure that the model in \Cref{eq: gp-conditional-mean-1} satisfies the PDE at $n_{PDE}$ randomly selected collocation points (CPs) in the domain. 

\subsection{Model Characteristics and Extensions} \label{sec: character-extension}
In this section, we elaborate on four unique features of our approach for solving PDE systems and discuss two major extensions that enable $(1)$ data fusion for solving inverse problems, and $(2)$ efficiently solving PDEs with multiple dependent variables. 
We highlight that combining GPs and NNs was first introduced in \cite{RN1919} to improve the uncertainty quantification power of NNs in supervised learning tasks. 
However, in sharp contrast to our work, the proposed approach in \cite{RN1919} relies on big data in the entire domain (we do not use any labeled data in the domain), aims to improve prediction uncertainties, leverages MLE for parameter optimization, and hinges on sparse GPs for scalability. 

\noindent \textbf{Feature one:} The overall training cost of our approach almost entirely depends on the second module since selecting $\widehat\phib$ does not rely on MLE and is a computationally inexpensive process. 
Additionally, our experiments consistently indicate that the performance of the final model across a broad range of problems is quite robust to $\widehat\phib$ as long as the BCs/IC are sufficiently sampled (see \Cref{sec: sensitivity-analyses} for sample results). This robustness is independent of the random values assigned to $\thetab$ in module $1$. Based on these two observations, in all of our experiments we simply assign $10^2$ to all the length-scale parameters of the kernel and sample $40$ points at each boundary. We highlight that while these values are certainly not the optimum, they have consistently enabled us to achieve highly competitive results. 

\begin{figure}[!t]
    \centering
    \includegraphics[width=1\textwidth]{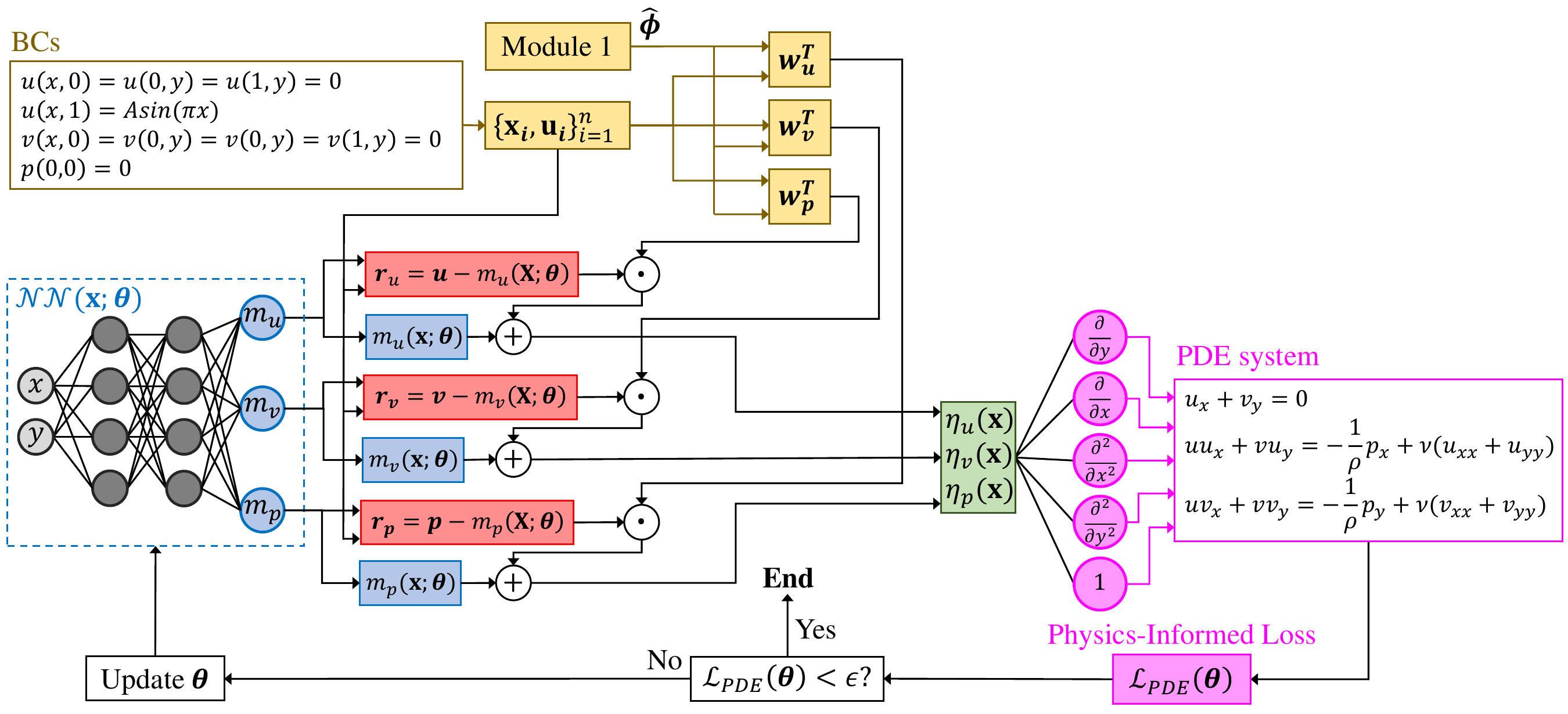}
    \caption{\textbf{Solving the 2D incompressible Navier-Stokes equations for the lid-driven cavity problem:} With minor architectural changes on module two with respect to \Cref{fig: flowchart}, our framework can also solve coupled PDE systems. Specifically, we endow each dependent variable with a GP prior. These GPs have independent kernels but a shared mean function that is parameterized via a deep neural network. Similar to \Cref{fig: flowchart}, the loss function only depends on the PDE residuals and excludes data loss terms on BC/IC.}
    \label{fig: ldc-flowchart}
\end{figure}

\noindent \textbf{Feature two:} The computational cost of coupling GPs and deep NNs in our framework is negligible during both training and testing since $\Cb$ does not change in the second module and its size only depends on $\outputub$ which is a relatively small vector. When solving PDEs such as the Navier-Stokes equations that have multiple dependent variables, the size of $\outputub$ can grow rapidly since it will store boundary and initial data on multiple outputs. Hence, for such PDEs we decouple the kernel-weighted \shortspace of the outputs to keep the size of $\Cb$ and $\outputub$ small. As shown in \Cref{fig: ldc-flowchart}, we formulate this decoupling by endowing the dependent variables with a collection of GP priors which share the same mean function but have independent kernels (i.e., the predictive formula in \Cref{eq: gp-conditional-mean-1} is used for as many outputs as the PDE has). 
While a single kernel can help in learning the inter-variable relations, we avoid this formulation for the following two main reasons. 
Firstly, it increases the size and condition number of the covariance matrix especially if the BCs on these variables are significantly different. For instance, on the top edge ($y=1$) in the lid-driven cavity (LDC) benchmark problem formulated in \Cref{eq: ldc}, pressure is unknown while the vertical and horizontal velocity components are equal to, respectively, zero and $A\sin{\parens{\pi x}}$. 
Secondly, our empirical findings indicate that the shared mean function is able to adequately learn the hidden interactions between these dependent variables. 

\begin{figure*}[!b]
    \centering
    \begin{subfigure}[t]{0.63\textwidth}
        \centering
        \includegraphics[width=1.00\columnwidth]{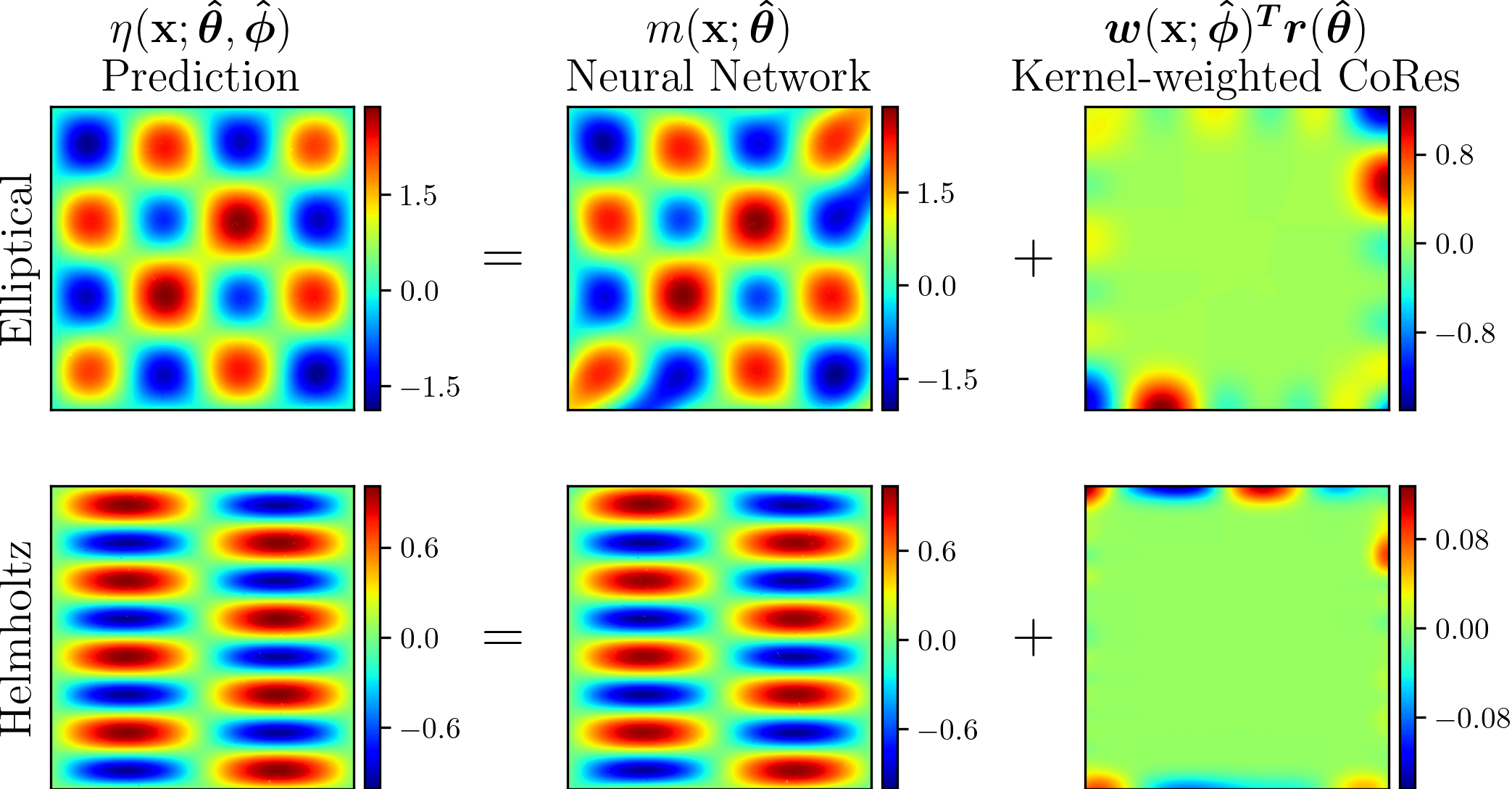}
        \captionsetup{justification=centering}
        \caption{Boundary value problems with direction-dependent solution frequencies.}
        \label{fig: CoRe}
    \end{subfigure}
    \begin{subfigure}[t]{0.34\textwidth}
        \centering
        \includegraphics[width=1.00\columnwidth]{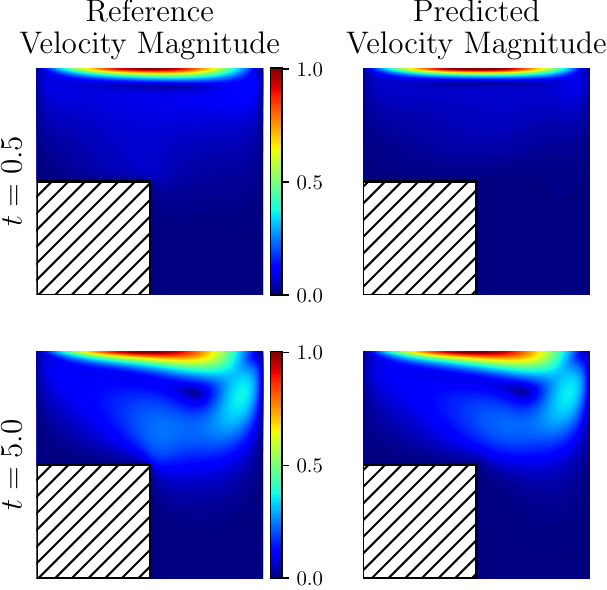}
        \captionsetup{justification=centering}
        \caption{Time-dependent LDC problem.}
        \label{fig: l-shape-ns-xyt}
    \end{subfigure}%
    \vspace{2mm}
    \caption{\textbf{Model features: (a)} In addition to improving the optimization of $\thetab$ in module two, kernel-weighted \shortspace contribute to the model predictions and ensure strict satisfaction of the BCs. \textbf{(b)} Kernel-weighted \shortspace automatically adapt to the domain geometry and are applicable to coupled PDE systems such as the Navier-Stokes equations. Here, we solve the unsteady LDC problem for $t \in \brackets{0, 5}$ and visualize the flow at $t=0.5$ and $t=5$.}
    \label{fig: model-features}
\end{figure*}

\noindent \textbf{Feature three:} As proven in \Cref{sec: proof}, our model can exactly satisfy the BCs/IC as the number of sampled boundary points increases. Due to this feature, the loss function in \Cref{fig: flowchart} or \Cref{fig: ldc-flowchart} only minimizes the error in satisfying the PDE and excludes data loss terms that encourage the NN to reproduce the BCs/IC. This exclusion indicates that our framework does not need weight balancing which is an expensive process that ensures each component of a composite loss is appropriately minimized during training. 
We highlight that per \Cref{eq: gp-conditional-mean} the contribution of kernel-weighted \shortspace to the model's predictions decreases as the distance with the boundaries (where the data is available) increases. As shown in \Cref{fig: CoRe} this decrease is not sudden and depends on the quality of the learnt mean function.

The proof in \Cref{sec: proof} indicates that the convergence as $n\rightarrow\infty$ is independent of the domain geometry and the potential noise that may corrupt the boundary data. The former feature is especially useful when solving PDEs over irregular domains such as the unsteady LDC problem in \Cref{fig: l-shape-ns-xyt}.

The above three features imply that training a PINN (or any of its extensions) costs similarly to the case where the same network is used as $m(\inputb; \thetab)$ in our framework. This behavior is very attractive since our approach consistently and substantially improves the performance of existing NN-based methods while also simplifying the training process by dispensing with the need for balancing the loss terms or fine-tuning the optimization settings.

\noindent \textbf{Feature four:} Our framework allows to perform data fusion and system identification by incorporating the additional measurements or observations in the kernel structure in exactly the same way that BC/IC data are handled, see \Cref{sec: inverse-problems} for a graphical flowchart and some examples (note that since we handle observations similarly to BC/IC data, the reproducibility proof in \Cref{sec: proof} applies to them as well). 
This extension provides fast convergence rates for solving inverse problems and enables combining multiple data sets (e.g., experiments and simulations) to discover missing physics or unknown PDE parameters.

    \section{Results and Discussions}\label{sec: results}
We compare the performance of our approach against four baseline PIML methods on four different PDE systems. 
We describe these PDE systems in \Cref{sec: pde-description} and then provide some details on implementation and training/testing of the five PIML models in \Cref{sec: implementation-details} where we also comment on our rationale for choosing the baselines. 
We summarize the results of our comparative studies in \Cref{sec: comparative-study} and then conduct extensive sensitivity analyses in \Cref{sec: sensitivity-analyses} to assess the impact of factors such as random initialization, noise, network architecture, and optimization settings on the summary results reported in \Cref{sec: comparative-study}. We conclude this section by comparing the training loss trajectory of our model to those of PINNs in \Cref{sec: loss-behavior}. Additional experiments supporting our claims are provided in \Cref{sec: additionalResults}.



\subsection{Description of the Benchmark Problems} \label{sec: pde-description}
The four PDE systems used in our studies are described below. Each problem is solved under two settings to understand the effect of PDE complexity on the performance of the PIML models.  

Throughout each problem, we specify the collection of independent variables $\inputb$, \eg $\inputb \coloneq \brackets{x,t}$ in Burgers' equation. In the case of Navier-Stokes equation, the collection of dependent variables is denoted by $\outputub(\inputb) \coloneq \brackets{u(\inputb), v(\inputb), p(\inputb)}^T$.

\noindent \textbf{Burgers' Equation:}
We consider a viscous system subject to IC and Dirichlet BC in one space dimension:
\begin{equation}
    \begin{aligned}
        &\ut + u \ux - \nu \uxx = 0, && \forall x \in (-1,1), t \in (0, 1] \\
        &u(-1, t) = u(1, t) = 0, && \forall t \in [0, 1] \\
        &u(x, 0) = -\sin{\parens{\pi x}}, && \forall x \in [-1,1]
    \end{aligned}
    \label{eq: Burgers-main}
\end{equation}
where $\inputb \coloneq \brackets{x, t}$ and $\nu$ is the kinematic viscosity. \Cref{eq: Burgers-main} frequently arises in fluid mechanics and nonlinear acoustics.
In our studies, we investigate the performance of different PIML models in solving \Cref{eq: Burgers-main} for $\nu \in \braces{\frac{0.01}{\pi}, \frac{0.02}{\pi}}$ which controls the solution smoothness at $x=0$ where a shock wave forms as $\nu$ approaches zero.  To broaden the range of PDEs that our proposed method can address, we also consider the inviscid Burgers' equation ($\nu = 0$) which has a shock in \Cref{sec inviscid burgers}.

\noindent \textbf{Nonlinear Elliptic PDE:}
To assess the ability of our approach in learning high-frequency solutions, we study the boundary value problem developed in \cite{RN1886}:
\begin{equation}
    \begin{aligned}
        &\uxx + \uyy - \alpha u^3 = f\parens{x,y}, && \forall x,y \in (0,1)^2 \\
        &u(x, 0) = u(x, 1) = 0, && \forall x \in [0, 1] \\
        &u(0, y) = u(1, y) = 0, && \forall y \in [0, 1]
    \end{aligned}
    \label{eq: elliptic}
\end{equation}
where $\inputb \coloneq \brackets{x, y}$ and $\alpha \in \braces{20, 30}$ is a constant that controls the nonlinearity degree. 
$f(x,y)$ is designed such that the solution is $u(x,y) = \sin{\parens{\pi x}} \sin{\parens{\pi y}} + 2 \sin{\parens{4 \pi x}} \sin{\parens{4\pi y}}$.

\noindent \textbf{Eikonal Equation:}
We consider the two-dimensional regularized Eikonal equation \cite{RN1886} which is typically encountered in the context of wave propagation:
\begin{equation}
    \begin{aligned}
        &\ux^2 + \uy^2 -\epsilon \parens{\uxx + \uyy} = 1, && \forall x, y \in (0,1)^2 \\
        &u(x, 0) = u(x, 1) = 0, && \forall x \in [0, 1] \\
        &u(0, y) = u(1, y) = 0, && \forall y \in [0, 1]
    \end{aligned}
    \label{eq: eikonal}
\end{equation}
where $\inputb \coloneq \brackets{x, y}$ and $\epsilon \in \braces{0.01, 0.05}$ is a constant that controls the smoothing effect of the regularization term.

\noindent \textbf{Lid-Driven Cavity (LDC):}
The two-dimensional steady state LDC problem has become a gold standard for evaluating the ability of PIML models in solving coupled PDEs. 
This problem is governed by the incompressible Navier-Stokes equations:
\begin{equation}
    \begin{aligned}
        &\ux + \vy = 0, && \forall \inputb \in (0,1)^2 \\
        &u \ux + v \uy = -\frac{1}{\rho}\px + \nu \parens{\uxx + \uyy}, && \forall \inputb \in (0,1)^2 \\
        &u \vx + v \vy = -\frac{1}{\rho}\py + \nu \parens{\vxx + \vyy}, && \forall \inputb \in (0,1)^2 \\
        &v(x, 0) = v(x, 1) = v(0, y) = v(1, y) = 0, && \forall x,y \in [0,1] \\
        &u(x, 0) = u(0, y) = u(1, y) = 0, && \forall x,y \in [0,1] \\
        &u(x, 1) = A\sin{\parens{\pi x}}, && \forall x \in [0, 1] \\      
        &p(0,0) = 0
    \end{aligned}    
    \label{eq: ldc}
\end{equation}
where $\inputb \coloneq \brackets{x, y}$, $\nu = 0.01$ is the kinematic viscosity, $\rho = 1.0$ denotes the density, and $A \in \braces{3, 5}$ is a scaling constant. 
The Reynolds number for this LDC problem can be computed via $\Rey = \frac{\rho \bar{u} L}{\nu}$ where $\bar{u} = \int_0^1 A\sin{\parens{\pi x}}dx$ is the characteristic speed of the flow and $L = 1$ is the characteristic length. For the two cases $A \in \braces{3, 5}$, we obtain $\Rey \in \braces{191, 318}$.

\subsection{Implementation Details in Our Comparative Studies} \label{sec: implementation-details}
Below, we first describe the architecture and training procedure of the PIML models used throughout our paper and then comment on how the reference solutions are obtained for each PDE system. 

\subsubsection{Architecture and Training}
We use a fully connected feed-forward NN as the mean function in our framework and design its input and output dimensionality based on the PDE system. We denote our model via NN-\shortspace and compare it against $(1)$ \GPorSpace which is the optimal recovery approach of \cite{RN1886} that leverages zero-mean GPs, $(2)$ PINNs whose architectures are exactly the same as our NNs in the mean function $m\parens{\inputb; \thetab}$, $(3)$ \PINNdw~which is a variation of PINNs that balances loss components with dynamic weights \cite{wang2021understanding}, and $(4)$ \PINNhc~which is a PINN whose output is designed to strictly satisfy the BCs/IC \cite{berg2018unified}. More detailed information about these four models is provided in \Cref{sec: methods-description}. 

Our rationale for comparing our method against \GPor, PINNs, \PINNdw, and \PINNhc~are as follows. 
Evaluation against \GPorSpace assesses our arguments on the limitations of using zero-mean GPs for solving PDEs when labeled solution data are only available as IC/BC. Comparisons against vanilla PINNs directly show the impact of kernel-weighted \shortspace as the same network architecture is used as the mean function in NN-\short. Lastly, comparisons against \PINNdw~and \PINNhc~highlight the benefits of NN-\shortspace in automatically satisfying the BC/IC.

The NN-based approaches (i.e., NN-\short, PINN, \PINNdw,  and \PINNhc) are all implemented in PyTorch \cite{paszke2019pytorch} and use hyperbolic tangent activation functions in all their layers except the output one where a linear activation function is used. The number and size of the hidden layers are exactly the same across these methods to enable a fair and straightforward comparison. 
For NN-\shortspace we use the Gaussian kernel in \Cref{eq: kernel-main2} with $\sigma^2=1$ and $\phib = 10^{\omegab}=10^{\bm{2}}$ in all the simulations in \Cref{sec: comparative-study} (note that, all the length-scale parameters are fixed to $10^2$, we study the effect of other values in \Cref{sec: sensitivity-analyses}). The nugget or jitter parameter of the kernel in \Cref{eq: kernel-main2} is chosen such that the covariance matrix is numerically stable. We ensure this stability by imposing an upper bound of approximately $\kappa_{max} \approx10^6$ on the condition number of the covariance matrix, i.e., $\kappa<\kappa_{max}$. This constraint typically results in a nugget value of around $10^{-5}$ or $10^{-4}$. We have not optimized the performance of NN-\shortspace with respect to $\kappa_{max}$ as we have found our current results to be sufficiently accurate. 

To optimize NN-\short, PINNs, and \PINNhc \ we leverage L-BFGS with a learning rate of $10^{-2}$ while \PINNdw \ is optimized using Adam with a learning rate of $10^{-3}$ (note that the performance of L-BFGS deteriorates if dynamic weights are used in the loss function). 
To ensure these NN-based methods produce optimum models, we use a very large number of epochs during training. Specifically, we employ $1{,}000$ and $2{,}000$ epochs for single- and multi-output problems, respectively. Since Adam typically requires more epochs for convergence, we train \PINNdw \ for $40{,}000$ epochs across all problems. 
To evaluate the loss function, we use $10{,}000$ collocation points within the domain in all cases. For PINN and \PINNdw~we uniformly sample boundary and/or initial conditions at $1{,}000$ locations while we only sample $40$ points for NN-\short. This significant difference is due to the fact that we observed that NN-\shortspace with just $40$ boundary points can outperform other methods. Leveraging more boundary data improves the performance of NN-\shortspace in solving PDE systems especially in satisfying the IC/BC. 

We fit \GPorSpace based on the code and specifications provided by \cite{RN1886} which leverages a variant of the Gauss–Newton algorithm for optimization. The performance of \GPorSpace depends on the kernel parameters and the number of interior nodes $n_{PDE}$ where $\zb$ needs to be estimated. For the former, we use the recommended values in \cite{RN1886} and for the latter we choose two values ($1{,}000$ and $2{,}000$) in our experiments. 

NN-\short, PINN, \PINNdw, and \PINNhc \ are trained on an NVIDIA GeForce RTX 3060 with 64 GB of RAM whereas \GPorSpace is trained on a CPU equipped with a 11th Gen Intel-Core i7-11700K running at a base clock speed of $3.6$ GHz. The training cost of NN-\shortspace compared to PINN for each problem is reported in \Cref{table: time-comparison} and discussed in \Cref{sec: comp-cost}.

\subsubsection{Reference Solutions and Accuracy Metric}
We obtain the reference solutions for the PDE systems as follows:
\begin{itemize}
    \item Burgers' Equation: The reference solution is obtained from the code provided in \cite{RN1886} which employs the Cole-Hopf transformation \cite{ohwada2009cole} together with the numerical quadrature. 
    \item Elliptic PDE: The analytical solution for this problem is $u(x,y) = \sin{\parens{\pi x}} \sin{\parens{\pi y}} + 2 \sin{\parens{4 \pi x}} \sin{\parens{\pi y}}$.
    \item Eikonal Equation: We leverage the solution method provided by \cite{RN1886} which applies the transformation $u(x,y) = -\epsilon \log{g(x,y)}$ leading to the linear PDE $g - \epsilon^2 \Delta g = 0$ that can be solved via the finite difference method. 
    \item Lid-Driven Cavity: we use the finite element method implemented in the commercial software package COMSOL \cite{multiphysics1998introduction}. 
\end{itemize}

To quantify the accuracy of the PIML models, we calculate the Euclidean norm of the error between the reference and predicted solutions at $n_t=10^4$ randomly chosen points. We denote this error metric via $L_{2,e}$ and calculate it as:
\begin{equation}
    L_{2,e} =  \sqrt{\frac{1}{n_t} \sum_{i=1}^{n_t} \parens{\outputu(\inputb_i) - \eta(\inputb_i;\widehat\thetab, \widehat\phib)}^2}.
    \label{eq: l2-error}
\end{equation}

\subsection{Summary of Comparative Studies} \label{sec: comparative-study}

\begin{figure*}[!t]
    \centering
    \includegraphics[width=1.0\textwidth]{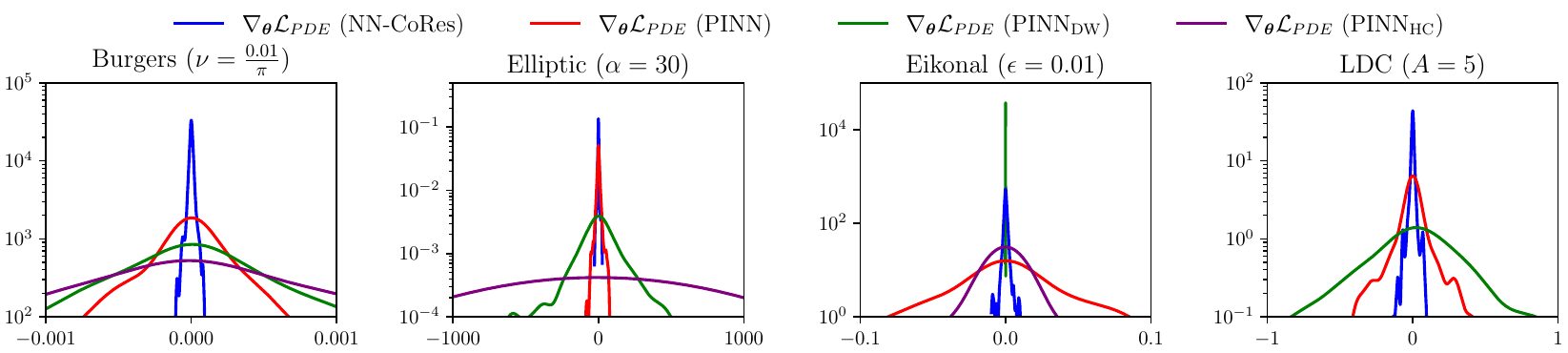}
    \caption{\textbf{Histograms of PDE loss gradients:} NN-\shortspace is in general more effective in minimizing its loss function as a larger portion of its gradients satisfy the first order optimality condition. While \PINNdw~has more near-zero gradients in the Eikonal problem, it does so at the expense of violating the BC loss term. All models in this figure have a $4\otimes20$ architecture.} 
    \label{fig: loss-grad}
\end{figure*}

\begin{table*}[!b]
    \tiny
    \centering
    \caption{\textbf{Summary of comparative studies:} We report $L_{2,e}$ of different methods as a function of model capacity and PDE parameter. The symbol $\otimes$ indicates the network architecture (e.g., $4\otimes 10$ is an NN which has four $10-$ neuron hidden layers). Unlike NN-based methods, \GPor's accuracy relies on the number of interior nodes which we set to $1{,}000$ or $2{,}000$. \GPorSpace is not applied to LDC as it relies on manual derivation of the equivalent variational problem which, unlike the first three PDEs, is not done by the developers \cite{RN1886}. }
    \label{tab: comparison-main}
    \begin{tabular*}{\textwidth}{@{\extracolsep\fill}lccccccccccc}
        \toprule%
        & \multicolumn{2}{@{}c@{}}{\textbf{NN-\short}} & \multicolumn{2}{@{}c@{}}{\textbf{GP}$_\text{\textbf{OR}}$}  & \multicolumn{2}{@{}c@{}}{\textbf{PINN}} & \multicolumn{2}{@{}c@{}}{\textbf{PINN}$_\text{\textbf{DW}}$} & \multicolumn{2}{@{}c@{}}{\textbf{PINN}$_\text{\textbf{HC}}$} \\
        \cmidrule{2-3}\cmidrule{4-5} \cmidrule{6-7} \cmidrule{8-9} \cmidrule{10-11}%
        \diagbox[width=10em]{\textbf{Problem}}{\textbf{Capacity}} & $4\otimes 10$ & $4\otimes 20$ & $1{,}000$ & $2{,}000$ & $4\otimes 10$ & $4\otimes 20$ & $4\otimes 10$ & $4\otimes 20$ & $4\otimes 10$ & $4\otimes 20$ \\
        \midrule
        Burgers' \Centerstack{$\nu = 0.02/\pi$\\$\nu = 0.01/\pi$}  & \Centerstack{$\mathbf{0.80\mathbf{e}{-3}}$\\$1.91\mathrm{e}{-3}$} & \Centerstack{$0.89\mathrm{e}{-3}$\\$\mathbf{1.29\mathbf{e}{-3}}$} & \Centerstack{$2.24\mathrm{e}{-1}$\\$1.69\mathrm{e}{-1}$} & \Centerstack{$1.69\mathrm{e}{-1}$\\$2.08\mathrm{e}{-1}$} & \Centerstack{$2.42\mathrm{e}{-3}$\\$4.26\mathrm{e}{-3}$} & \Centerstack{$1.50\mathrm{e}{-3}$\\$4.38\mathrm{e}{-3}$} & \Centerstack{$2.86\mathrm{e}{-3}$\\$1.93\mathrm{e}{-2}$} & \Centerstack{$3.18\mathrm{e}{-3}$\\$5.79\mathrm{e}{-3}$} & \Centerstack{$3.41\mathrm{e}{-1}$\\$3.65\mathrm{e}{-1}$} & \Centerstack{$3.36\mathrm{e}{-1}$\\$3.52\mathrm{e}{-1}$} \\ \hline
        Elliptic~ \Centerstack{$\alpha = 20$\\$\alpha = 30$} & \Centerstack{$4.50\mathrm{e}{-3}$\\$4.38\mathrm{e}{-3}$} & \Centerstack{$\mathbf{2.04\mathbf{e}{-3}}$\\$\mathbf{1.24\mathbf{e}{-3}}$} & \Centerstack{$2.44\mathrm{e}{-3}$\\$7.08\mathrm{e}{-3}$} & \Centerstack{$4.06\mathrm{e}{-3}$\\$6.55\mathrm{e}{-3}$} & \Centerstack{$6.55\mathrm{e}{-1}$\\$8.45\mathrm{e}{-1}$} & \Centerstack{$4.68\mathrm{e}{-1}$\\$5.55\mathrm{e}{-1}$} & \Centerstack{$2.97\mathrm{e}{-1}$\\$1.69\mathrm{e}{-1}$} & \Centerstack{$1.26\mathrm{e}{-1}$\\$1.19\mathrm{e}{-1}$} & \Centerstack{$6.35\mathrm{e}{-1}$\\$2.89\mathrm{e}{-1}$} & \Centerstack{$5.95\mathrm{e}{-1}$\\$6.53\mathrm{e}{-1}$} \\ \hline
        Eikonal~ \Centerstack{$\epsilon = 0.05$\\$\epsilon = 0.01$} & \Centerstack{$0.52\mathrm{e}{-3}$\\$\mathbf{4.60\mathbf{e}{-3}}$} & \Centerstack{$\mathbf{0.37\mathbf{e}{-3}}$\\$4.99\mathrm{e}{-3}$} & \Centerstack{$1.76\mathrm{e}{-1}$\\$2.18\mathrm{e}{-1}$} & \Centerstack{$1.25\mathrm{e}{-1}$\\$2.06\mathrm{e}{-1}$} & \Centerstack{$2.76\mathrm{e}{-3}$\\$6.41\mathrm{e}{-3}$} & \Centerstack{$2.19\mathrm{e}{-3}$\\$6.38\mathrm{e}{-3}$} & \Centerstack{$2.01\mathrm{e}{-3}$\\$5.03\mathrm{e}{-3}$} & \Centerstack{$1.54\mathrm{e}{-3}$\\$4.97\mathrm{e}{-3}$} & \Centerstack{$2.89\mathrm{e}{-1}$\\$2.91\mathrm{e}{-1}$} & \Centerstack{$2.88\mathrm{e}{-1}$\\$3.42\mathrm{e}{-1}$} \\ \hline
        LDC~~~~ \Centerstack{$A = 3$\\$A = 5$}  & \Centerstack{$1.86\mathrm{e}{-1}$\\$3.11\mathrm{e}{-1}$} & \Centerstack{$\mathbf{8.67\mathbf{e}{-2}}$\\$\mathbf{2.79\mathbf{e}{-1}}$} & \Centerstack{$-$\\$-$} & \Centerstack{$-$\\$-$} & \Centerstack{$2.72\mathrm{e}{-1}$\\$7.17\mathrm{e}{-1}$} & \Centerstack{$1.28\mathrm{e}{-1}$\\$6.77\mathrm{e}{-1}$} & \Centerstack{$3.01\mathrm{e}{-1}$\\$7.16\mathrm{e}{-1}$} & \Centerstack{$1.25\mathrm{e}{-1}$\\$6.23\mathrm{e}{-1}$} & \Centerstack{$4.32\mathrm{e}{-1}$\\$1.03\mathrm{e}{0}$} & \Centerstack{$5.29\mathrm{e}{-1}$\\$9.11\mathrm{e}{-1}$} \\ \bottomrule
    \end{tabular*}
\end{table*}
The results of our studies are summarized in \Cref{tab: comparison-main} and indicate that our approach consistently outperforms other methods by relatively large margins. Interestingly, in most cases even the small NN-\shortspace with four $10-$neuron hidden layers achieve lower errors than the high capacity version of the competing methods; indicating NN-\shortspace more effectively use their  networks' capacity to learn the PDE solution. 
To visually compare the efficiency in capacity utilization across different NN-based models, in \Cref{fig: loss-grad} we provide the histogram of the PDE loss gradients with respect to $\thetab$ at the end of training. For any model, most gradients in such a histogram should be ideally close to zero following the first-order necessary conditions of the Karush–Kuhn–Tucker theorem \cite{arora2004introduction}. 
We observe that NN-\shortspace achieve the most near-zero gradients \textit{while} satisfying the BCs/IC. In contrast, \PINNhc, which is also designed to automatically satisfy the BCs/IC, struggles to minimize the PDE loss (note that models such as PINNs and \PINNdw~which do not strictly satisfy BCs/IC, can achieve lower overall loss values and perhaps show better first-order necessary conditions at the expense of violating the BCs/IC).

\begin{figure*}[!t]
    \centering    
    \begin{subfigure}[t]{\textwidth}
        \centering
        \includegraphics[width=1.0\textwidth]{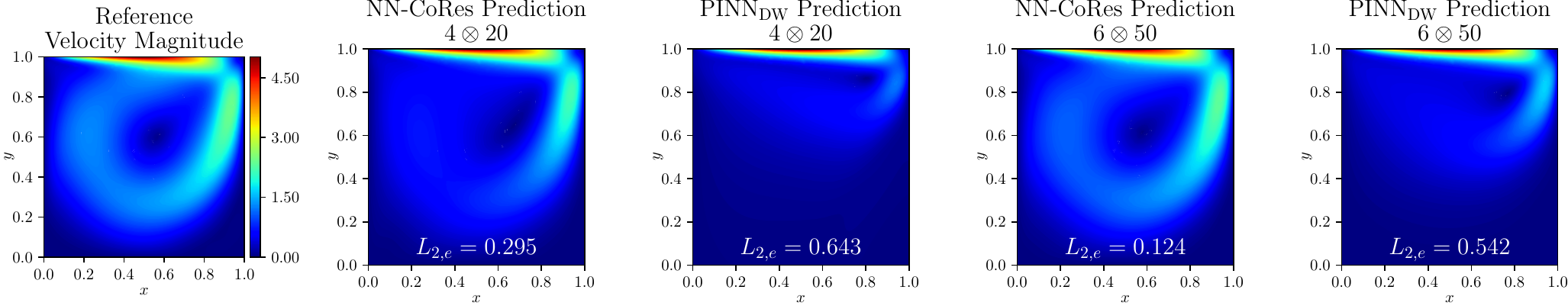}
        \caption{\textbf{Reference vs predictions (LDC with $A=5$):} Performance improves as the network sizes increase. The small NN-\shortspace is more accurate than the large \PINNdw.} 
        \label{fig: ldc-reference vs prediction}
    \end{subfigure}
    
    \vspace{\floatsep}  

    \begin{subfigure}[t]{\textwidth}
        \centering
        \includegraphics[width=1.0\textwidth]{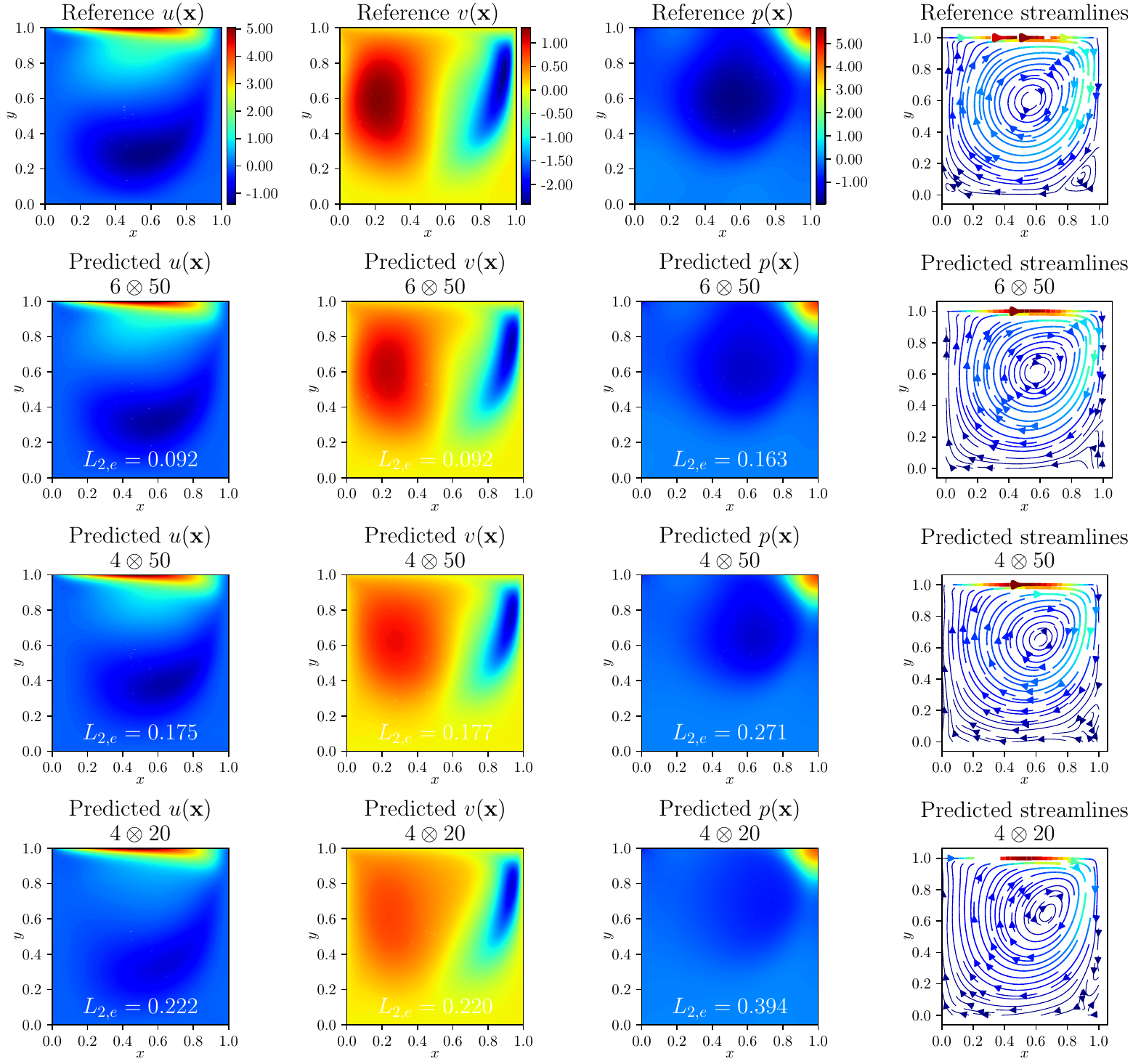}
        \caption{\textbf{Effect of network size on the accuracy in the LDC problem:} The accuracy of NN-\shortspace consistently increases in predicting $u, v,$ and $p$ as the network size (either the number of hidden layers or their size) increases.} 
        \label{fig: ldc-large-networks}
    \end{subfigure}
    
    \vspace{\floatsep}  
    
    \caption{\textbf{Error analysis:} NN-\shortspace achieve smaller errors and increasing the size of their networks provides more improvement compared to methods such as \PINNdw.}
    \label{fig: results}
\end{figure*}

We observe in \Cref{tab: comparison-main} that the performance of all the methods drops as either the problem complexity increases (e.g., Burgers' vs. LDC) or PDE parameters are changed to introduce more nonlinearity (e.g., $A=3$ vs $A=5$ in LDC).
This trend is expected since we do not change the architecture and training settings across our experiments. That is, we can increase the accuracy of all methods by increasing their capacity or improving the training process. 
We demonstrate this improvement for the LDC problem in \Cref{fig: ldc-reference vs prediction} where the errors of \PINNdw~and especially NN-\shortspace are decreased by merely increasing the size of their networks. To assess the convergence behavior of NN-\shortspace for this problem, we keep increasing the size of its mean function and observe in \Cref{fig: ldc-large-networks} that its performance improves (note that the training mechanism is not altered in these studies and merely the size of the models is increased). We also note that in the case of LDC problem pressure is only known at a single point on the boundary (rather than everywhere on the boundary) which indicates that the kernel-weighted \shortspace insignificantly help the model in learning pressure. We study this behavior further in \Cref{sec: loss-behavior}.

\begin{figure*}[t]
    \centering
    \includegraphics[width=1.0\textwidth]{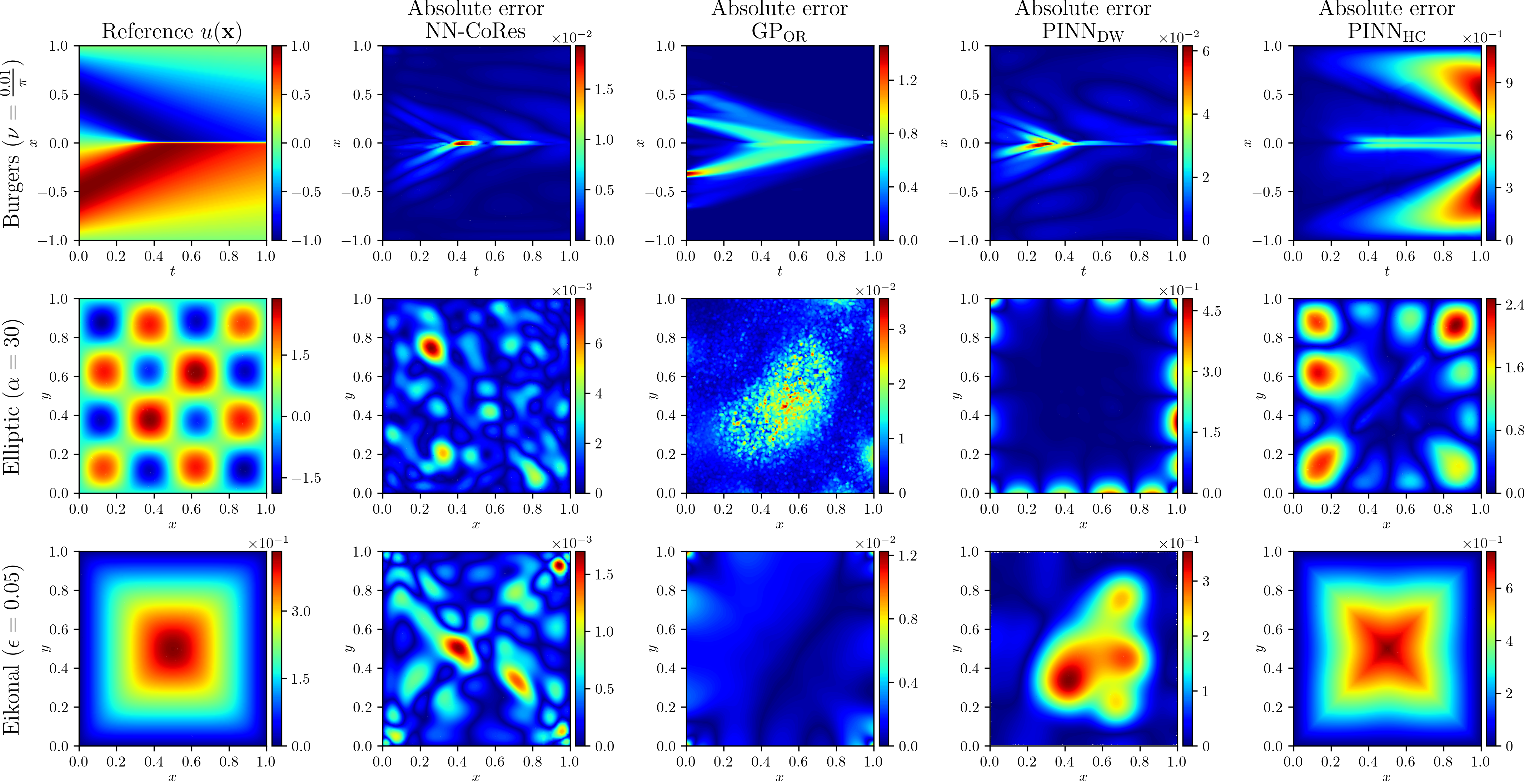}
    \caption{\textbf{Reference solutions and error maps:} NN-\shortspace provide much lower errors compared to other methods (note the scales of the error bars).}
    \label{fig: error-maps}
\end{figure*}

To gain more insight into the performance of each method, we visualize the error maps for some of our simulations in \Cref{fig: error-maps}.
We observe that \GPorSpace~is least accurate either in regions with sharp solution gradients or inside the domain where boundary information is not effectively propagated inward by the zero-mean GP. 
For \PINNdw, the errors are predominantly close to either the boundaries or where the PDE solution has large gradients. PINNs' errors in reproducing the BCs/IC are eliminated in \PINNhc~but at the expense of significant loss of accuracy elsewhere in the domain. 
These issues are largely addressed by NN-\shortspace which reproduce BCs/IC and approximate solutions with high gradients quite well. Specifically, we observe that in the case of Elliptic and Eikonal PDEs NN-\short' errors are quite evenly distributed in the domain and almost vanish close to the boundaries. In the case of Burgers' equation, NN-\short' errors are mostly around the $x=0$ line where the shock develops for an inviscid flow when $\nu=0$ (this error behavior is similar to other methods but the magnitude of the errors is much smaller for NN-\short).

\subsection{Sensitivity Analyses} \label{sec: sensitivity-analyses}
In this section, we conduct a wide range of sensitivity studies to assess the impact of factors such as random initialization, noise, network architecture, and optimization settings on the results reported in \Cref{sec: comparative-study}. 

\begin{figure*}[!t]
    \centering
    \begin{subfigure}[t]{\textwidth}
        \centering
        \includegraphics[width=1.0\textwidth]{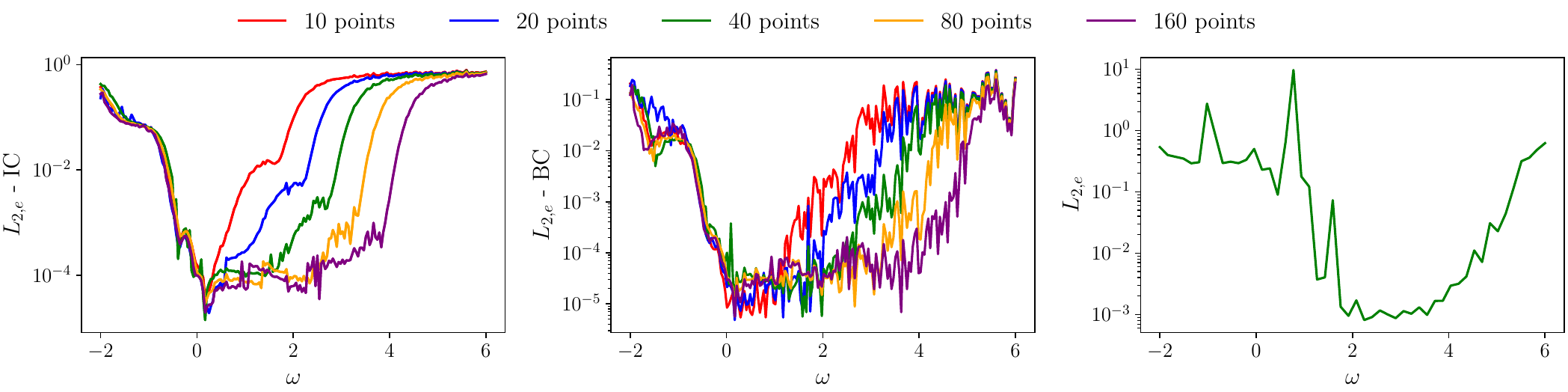}
        \captionsetup{justification=centering}
        \caption{Effect of $\omega$ on GP's interpolation power (left and middle plots) and NN-\shortspace (right plot). Burgers' equation is used in this study.}
        \label{fig: omega-effect}
    \end{subfigure}
    
    \vspace{\floatsep}  
    
    \begin{subfigure}[t]{\textwidth}
        \centering
        \includegraphics[width=1.0\textwidth]{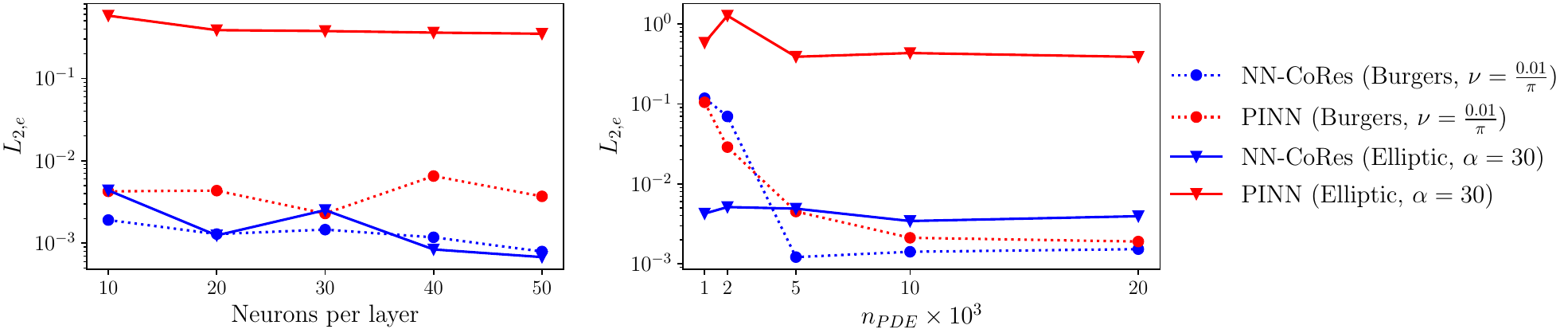}
        \captionsetup{justification=centering}
        \caption{Effect of network size (left, $n_{PDE}=10^4$) and $n_{PDE}$ (right, $4\otimes 20$ architecture) on the accuracy of PINNs and NN-\short.}
        \label{fig: nn-cp-effect}
    \end{subfigure}

     \vspace{\floatsep}  
    
    \begin{subfigure}[t]{\textwidth}
        \centering
        \includegraphics[width=0.8\textwidth]{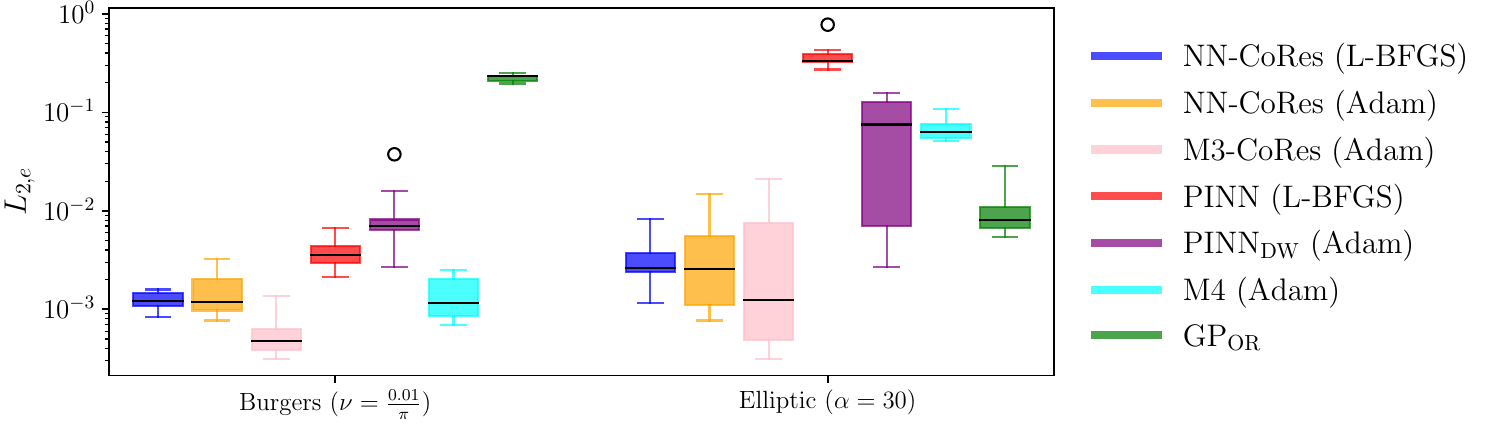}
        \captionsetup{justification=centering}
        \caption{Effect of optimizer, random initialization, and architecture type on errors for the Burgers' and Elliptic problems.}
        \label{fig: optimizer-initialization-arch-effect}
    \end{subfigure}

    \vspace{\floatsep}

    \begin{subfigure}[t]{\textwidth}
        \centering
        \includegraphics[width=1.0\textwidth]{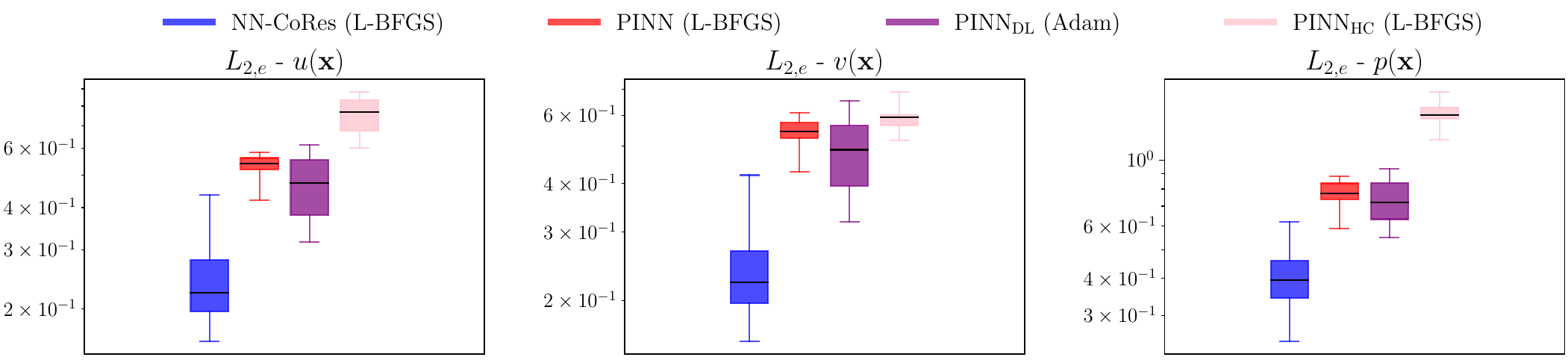}
        \captionsetup{justification=centering}
        \caption{Effect of random initialization and optimizer on errors for the LDC problem ($A=5$). All models have $4\otimes20$ architecture and use $n_{PDE}=10^4$.}
        \label{fig: initialization-effect-ldc}
    \end{subfigure}
    
    \caption{\textbf{Sensitivity studies:} We analyze the sensitivity of our results to factors such as the roughness parameters in the kernel, optimization settings, network architecture, and initialization. Based on these experiments, NN-\shortspace provide a more robust machine learning-based approach for solving different nonlinear PDEs.}
    \label{fig: sensitivity}
\end{figure*}

As stated in \Cref{sec: methodology-all}, the performance of NN-\shortspace is quite robust to the values chosen for $\phib=10^{\omegab}$ as long as they lie within a certain range. To obtain this range, we conduct the following inexpensive experiment using the Burgers' problem in \Cref{eq: Burgers-main} and $c\parens{\inputb, \inputb'; \phi, \delta, \sigma^2=1} =  \exp\braces{-\phi\parens{x-x'}^2 - \phi\parens{t-t'}^2} + \mathbbm{1}\{\inputb==\inputb'\}\delta$. 
We first sample $n_{train}$ equally spaced boundary samples using the provided analytic IC and BCs. To quantify the effect of data size on the results, we consider $5$ scenarios where $n_{train} \in \{10, 20, 40, 80, 160\}$. For each of these five cases, we build $200$ independent GPs whose only difference is the value that we assign to $\phi=10^{\omega}$. Specifically, we consider $200$ equally spaced values in the $[-2, 6]$ range for $\omega$ and use each of these values in one of the GPs which all have a non-zero mean function (we use a deep NN whose parameters are randomly initialized and frozen as the mean function). Once these GPs are built, we use them to predict on $n_{test}=10^4$ boundary points (see \Cref{eq: gp-conditional-mean} for the prediction formula). The results of this study are shown in the left and middle plots in \Cref{fig: omega-effect} and indicate that as more training data are sampled on the boundaries a wider range of values for $\omega$ result in small test errors. We highlight that this study is computationally very fast since none of the GPs are optimized; rather their parameters are either chosen by us (i.e., $\omega$), or fixed (i.e., $\delta, \sigma^2,$ and parameters of the NN mean).

Following the above study, we have decided to use $40$ boundary points in NN-\short. Based on the left and middle plots in \Cref{fig: omega-effect}, $\omega=2$ seems to be a good choice (but not the optimum one) for minimizing the error in reproducing the IC and BCs. To see the effect of this choice on the performance of a trained NN-\short, we again vary $\omega$ ($50$ equally spaced values in the $[-2, 6]$ range) but this time we train an NN-\shortspace model for each value of $\omega$. We evaluate the performance of these models in solving the Burgers' equation by reporting the Euclidean norm of the error $L_{2,e}$ at $n_{test} = 10^4$ points randomly located in the domain. The results are shown in the right plot in \Cref{fig: omega-effect} and indicate that although $\omega=2$ is not the optimum choice, it yields a model whose performance is close to optimal (the optimum model is achieved via an $\omega$ close to $3$).

We now conduct a few extensive experiments to study the effect of network size and optimization settings on the performance of various NN-based models. 
First, we fix everything and increase the number of neurons in each hidden layer from $10$ to $50$ (at increments of $10$) and solve the Burgers' and Elliptic PDEs via both NN-\shortspace and PINNs. We then repeat this experiment but this time we fix the architecture to $4\otimes20$ and incrementally increase $n_{PDE}$ from $10^3$ to $10^4$. The results of these two experiments are summarized in  \Cref{fig: nn-cp-effect} and indicate that NN-\shortspace is much less sensitive to the problem than PINNs which perform quite well on Burgers' but fail at accurately solving the Elliptic PDE that has direction-dependent frequency. We also observe that NN-\shortspace provide lower errors than PINNs in almost all simulations.

In our next experiment, we study the effects of optimizer (L-BFGS vs Adam), random initialization, and architecture type on the performance of various models. To this end, we again consider the Burgers' and Elliptic PDE systems and solve them with six NN-based methods and \GPor. For each case we repeat the training process of each model $10$ times to quantify the effect of random initialization on the models' solution accuracy. For these experiments, we also consider a new network architecture that we denote by M3 which is introduced in \cite{wang2021understanding} and aims to improve gradient flows by designing feed-forward networks with connections that resemble transformers \cite{vaswani2017attention}. In our framework, we replace the architecture that is used in all of our studies (which is a feed-forward neural network or an FFNN) with M3 and train the model with Adam (the resulting model is denoted by M3-\short). We also train another NN-based model denoted by M4 \cite{wang2021understanding} whose architecture is the same as M3 but leverages dynamic weights in its loss function. We highlight that the simulations that leverage M3 as their architecture have more parameters (and hence learning capacity) than cases where FFNNs are used so we expect M3-based simulations to provide lower errors. 

The results of these simulations are summarized in \Cref{fig: optimizer-initialization-arch-effect} and indicate that 
$(1)$ NN-\shortspace and \GPorSpace are less sensitive to random initializations compared to PINNs and their variations,
$(2)$ unlike other models, NN-\shortspace performs well in both PDE systems, i.e., our framework provides a more transferable method for solving PDEs via machine learning, and
$(3)$ architectures besides simple FFNNs (such as M3) can also be used in our framework to achieve higher accuracy.

The above experiments are based on the Burgers' and Elliptic PDE problems but our studies indicate that similar trends appear in other problems. To demonstrate this, we solve the LDC problem via four NN-based models that either use L-BFGS or Adam as their optimizer. We repeat the training process of each model $10$ times to assess the effect of random parameter initialization on each model's performance. The results are summarized with the boxplots in \Cref{fig: initialization-effect-ldc} and agree with our previous findings that indicate NN-\shortspace consistently outperform other methods. 

\begin{figure}[!t]
    \centering
    \includegraphics[width=\textwidth]{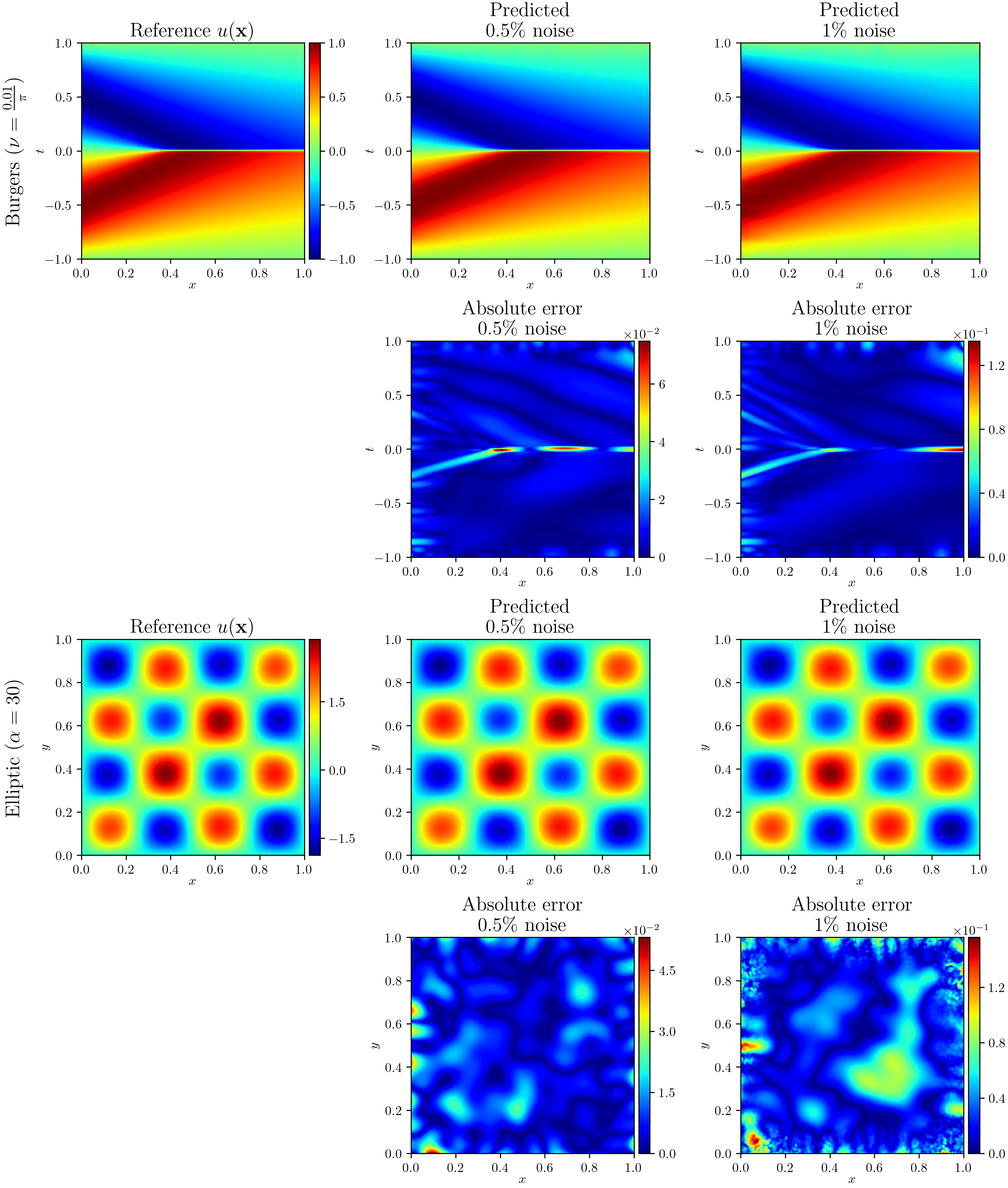}
    \caption{\textbf{Reference and predicted solutions with noisy boundary data:} We corrupt the samples obtained from boundary and initial conditions by either $0.5\%$ or $1\%$ of the solution range. In all cases, the performance of NN-\shortspace is insignificantly affected by the noise.}
    \label{fig: noisy-bc-ic}
\end{figure}

Finally, we investigate the effect of noisy boundary data on our results. Specifically, we corrupt the solution values that we sample from the IC and/or BCs before using them in our approach. We use a zero-mean normal distribution to model the noise and set the standard deviation to either $0.5\%$ or $1\%$ of the solution range. As shown in \Cref{fig: noisy-bc-ic}, the solution accuracy decreases as the noise variance increases (this trend is expected) but in all cases NN-\shortspace are able to quite effectively eliminate the noise and solve the Burgers' and Elliptic PDE systems.


\subsection{Loss Behavior} \label{sec: loss-behavior}
\begin{figure*}[!b]
    \centering
    \begin{subfigure}[t]{\textwidth}
        \centering
        \includegraphics[width=1.0\textwidth]{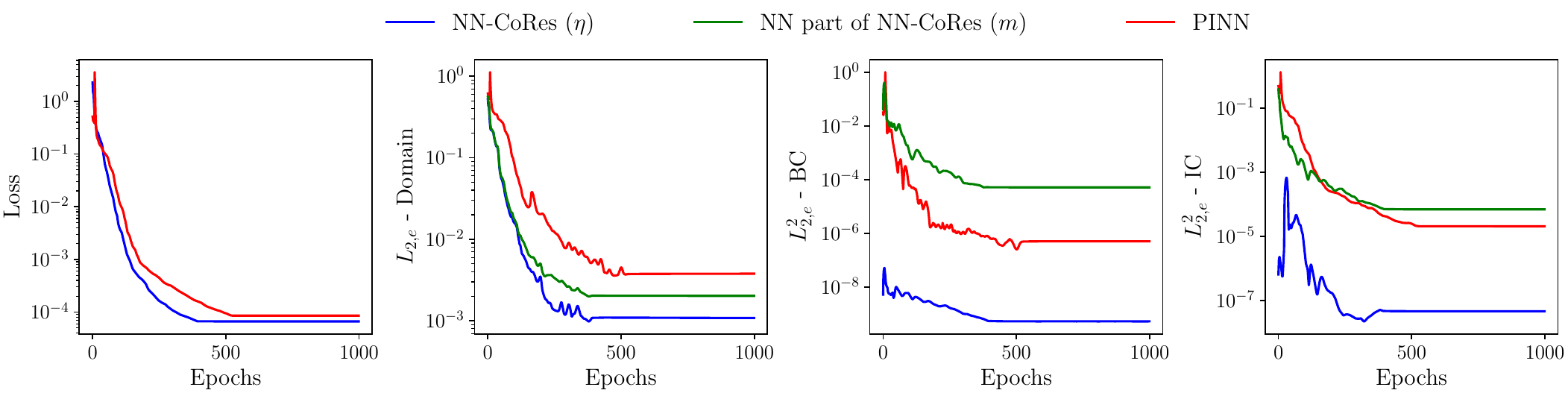}
        \captionsetup{justification=centering}
        \caption{Loss and accuracy history for Burgers' ($\nu = \frac{0.01}{\pi}$).}
        \label{fig: Burgers-loss-convergence-error-decomposition}
    \end{subfigure}
    
    \vspace{\floatsep}  
    
    \begin{subfigure}[t]{\textwidth}
        \centering
        \includegraphics[width=1.0\textwidth]{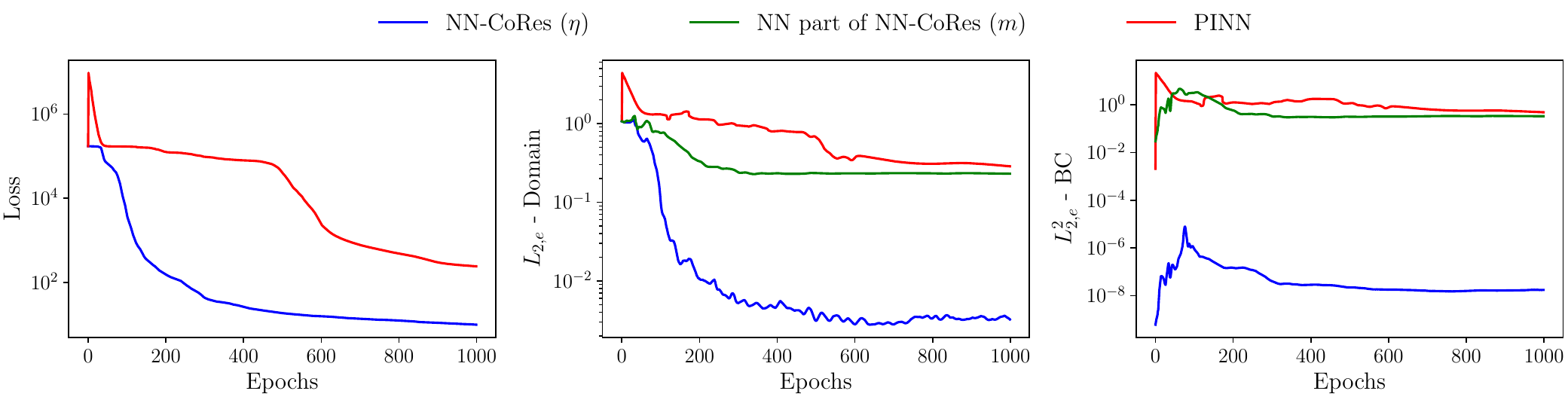}
        \captionsetup{justification=centering}
        \caption{Loss and accuracy history for Elliptic ($\alpha = 30$).}
        \label{fig: elliptic-loss-convergence-error-decomposition}
    \end{subfigure}

    \vspace{\floatsep}  
    
    \begin{subfigure}[t]{\textwidth}
        \centering
        \includegraphics[width=1.0\textwidth]{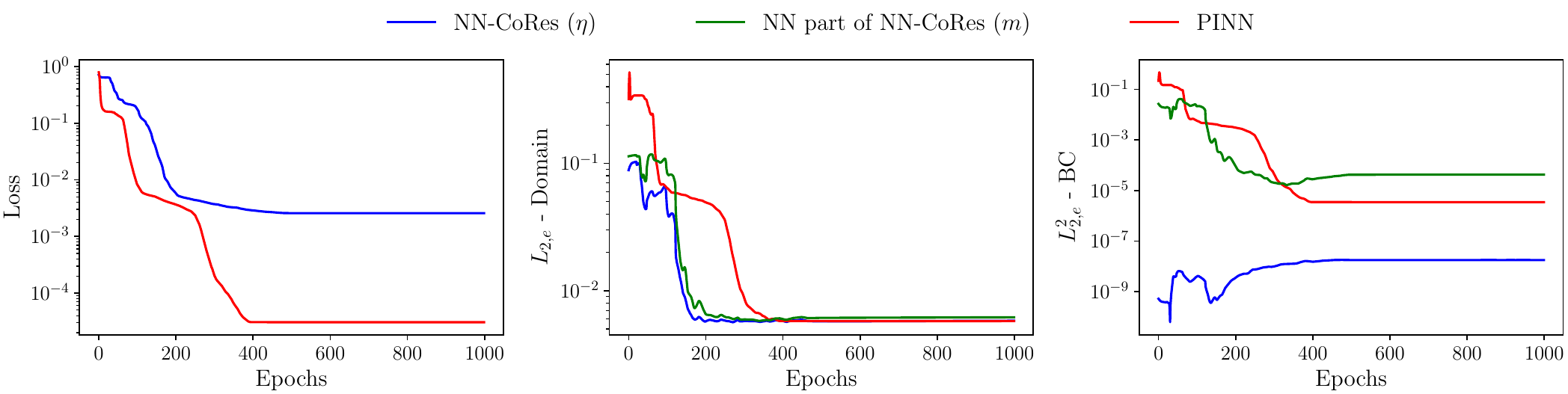}
        \captionsetup{justification=centering}
        \caption{Loss and accuracy history for Eikonal ($\epsilon = 0.01$).}
        \label{fig: eikonal-loss-convergence-error-decomposition}
    \end{subfigure}
    
    \caption{\textbf{Loss convergence and error decomposition:} NN-\shortspace typically converge faster than PINNs and consistently provide more accurate solutions. The NN part of NN-\shortspace benefits from the  kernel-weighted \shortspace not only on the boundaries, but also inside the domain.}
    \label{fig: loss-convergence-error-decomposition}
\end{figure*}
\begin{figure}[!t]
    \centering
    \includegraphics[width=\textwidth]{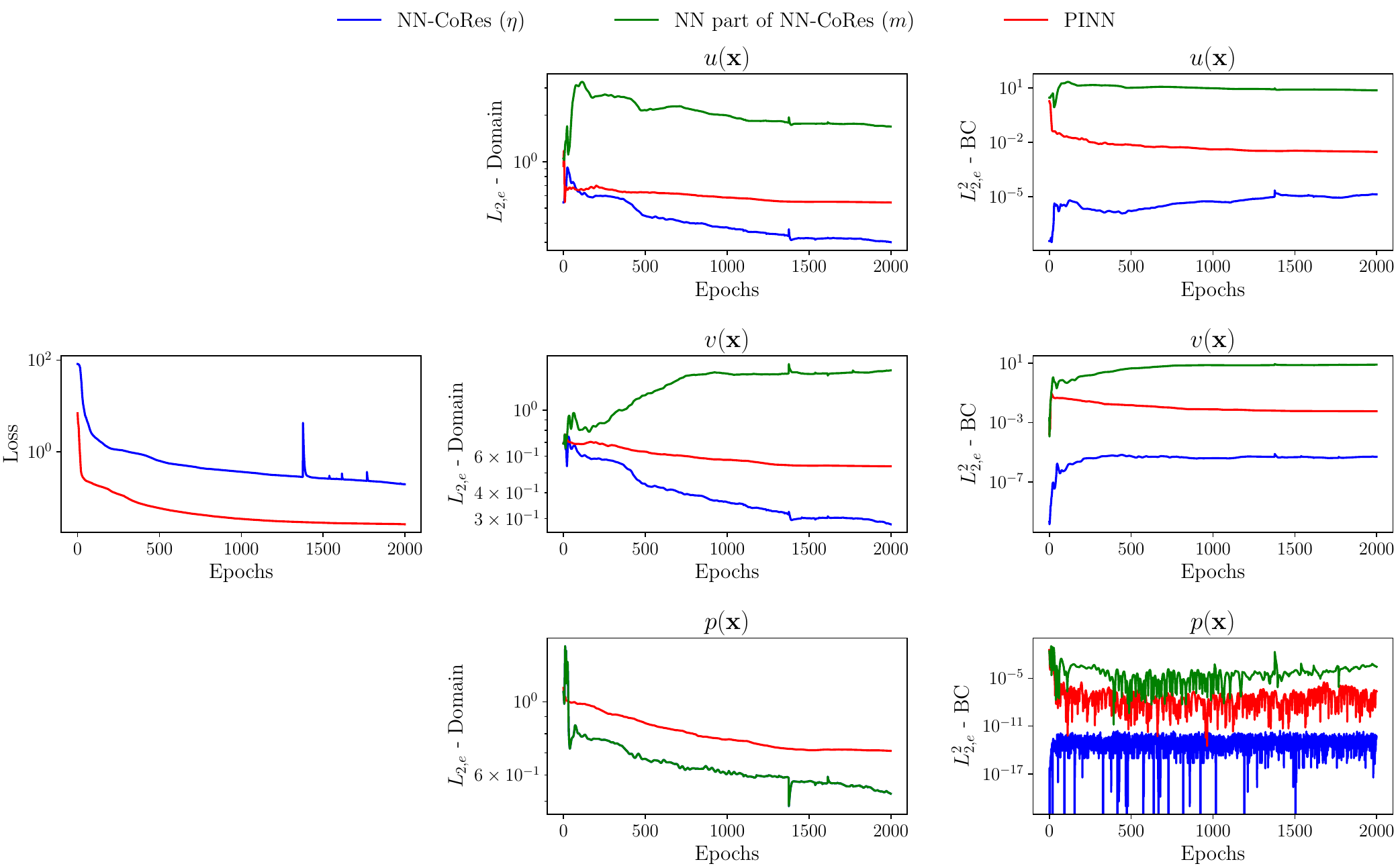}
    \caption{\textbf{Loss convergence and error decomposition for LDC:} The NN part of NN-\shortspace benefits from the  kernel-weighted \shortspace not only on the boundaries, but also inside the domain. In the case of pressure,  kernel-weighted \shortspace do not contribute to the model's predictions inside the domain as $p(\inputb)$ is only known at a single point on the boundary (for this reason, the blue and green curves are indistinguishable).}
    \label{fig: loss-convergence-error-decomposition-ldc}
\end{figure}

To gain more insights into the training dynamics of our approach, we visualize the loss and accuracy during the training process in \Cref{fig: loss-convergence-error-decomposition} and \Cref{fig: loss-convergence-error-decomposition-ldc} where in the latter figure we track the errors individually for each output for the LDC problem. We provide these plots for both PINNs and NN-\shortspace where the loss function of the former is based on \Cref{eq: pinn-loss} while NN-\shortspace only use $\mathcal{L}_{PDE}(\thetab)$ in their loss function. The solution accuracy is measured based on $L_{2,e}$ and $\parens{L_{2,e}}^2$ for points inside the domain and on its boundaries. Note that we square $L_{2,e}$ on the boundaries to be able to directly see its contribution to PINNs' loss, see $\mathcal{L}_{BC}(\thetab)$ and $\mathcal{L}_{IC}(\thetab)$ in \Cref{eq: pinn-loss}.
In the case of NN-\short, we also report the accuracy of its NN part on predicting the PDE solution to quantify the contributions of kernel-weighted \shortspace towards the model's predictions. 

As it can be observed in \Cref{fig: loss-convergence-error-decomposition,fig: loss-convergence-error-decomposition-ldc}, NN-\shortspace typically converge faster than PINNs, see the plots whose $y-$axis title is $L_{2,e}$ - Domain. We attribute this trend to the fact that, unlike in PINNs, the initial and boundary conditions are automatically satisfied in our models thanks to the kernel-weighted \shortspace which are smooth functions. This feature enables NN-\shortspace to focus on satisfying the PDE system in module two of our framework while PINNs have to struggle with both the differential equations as well as the initial and boundary conditions. 

An interesting trend in \Cref{fig: loss-convergence-error-decomposition,fig: loss-convergence-error-decomposition-ldc} is that the errors of NN-\shortspace are consistently lower than their NN components both in the domain and on the boundaries. That is, the kernel-weighted \shortspace positively contribute to the model's predictions both on the boundaries and inside the domain. This behavior is in sharp contrast to most approaches such as \PINNhc~that satisfy the boundary conditions at the expense of complicating the training process. 

Another interesting trend that we observe in \Cref{fig: loss-convergence-error-decomposition,fig: loss-convergence-error-decomposition-ldc} is that PINNs achieve lower loss values than NN-\shortspace in the case of Eikonal and LDC problems. While lower loss values are desirable, in these cases the observed trends are misleading. To explain this behavior, we note that the loss function of NN-\shortspace is simply $\mathcal{L}_{PDE}(\thetab)$ as the boundary and initial conditions are automatically satisfied. However, the loss function of PINNs minimizes $\mathcal{L}_{IC}(\thetab)$ and/or $\mathcal{L}_{BC}(\thetab)$ in addition to $\mathcal{L}_{PDE}(\thetab)$. That is, since PINNs do not strictly satisfy the BCs/IC, they are less regularized and hence can minimize $\mathcal{L}_{PDE}(\thetab)$ (which dominates the overall loss) in a more flexible manner. However, this behavior provides less accuracy since the boundary conditions are not learnt sufficiently well.

\subsection{Inverse Problems} \label{sec: inverse-problems}
In the previous experiments we have only used the differential equations along with the IC and/or BCs in building NN-\short. In this section, we introduce an extension of our framework for solving inverse problems where $(1)$ there are some (possibly noisy) labeled data available inside the domain (we refer to these samples as observations to distinguish them from the boundary data), and $(2)$ one or more parameters in the differential equations are unknown. Our goal in such applications is to solve the PDE system while estimating the unknown parameters. 

\begin{figure*}[!ht]
    \centering    
    
    \begin{subfigure}[t]{\textwidth}
        \centering
        \includegraphics[width=1.0\textwidth]{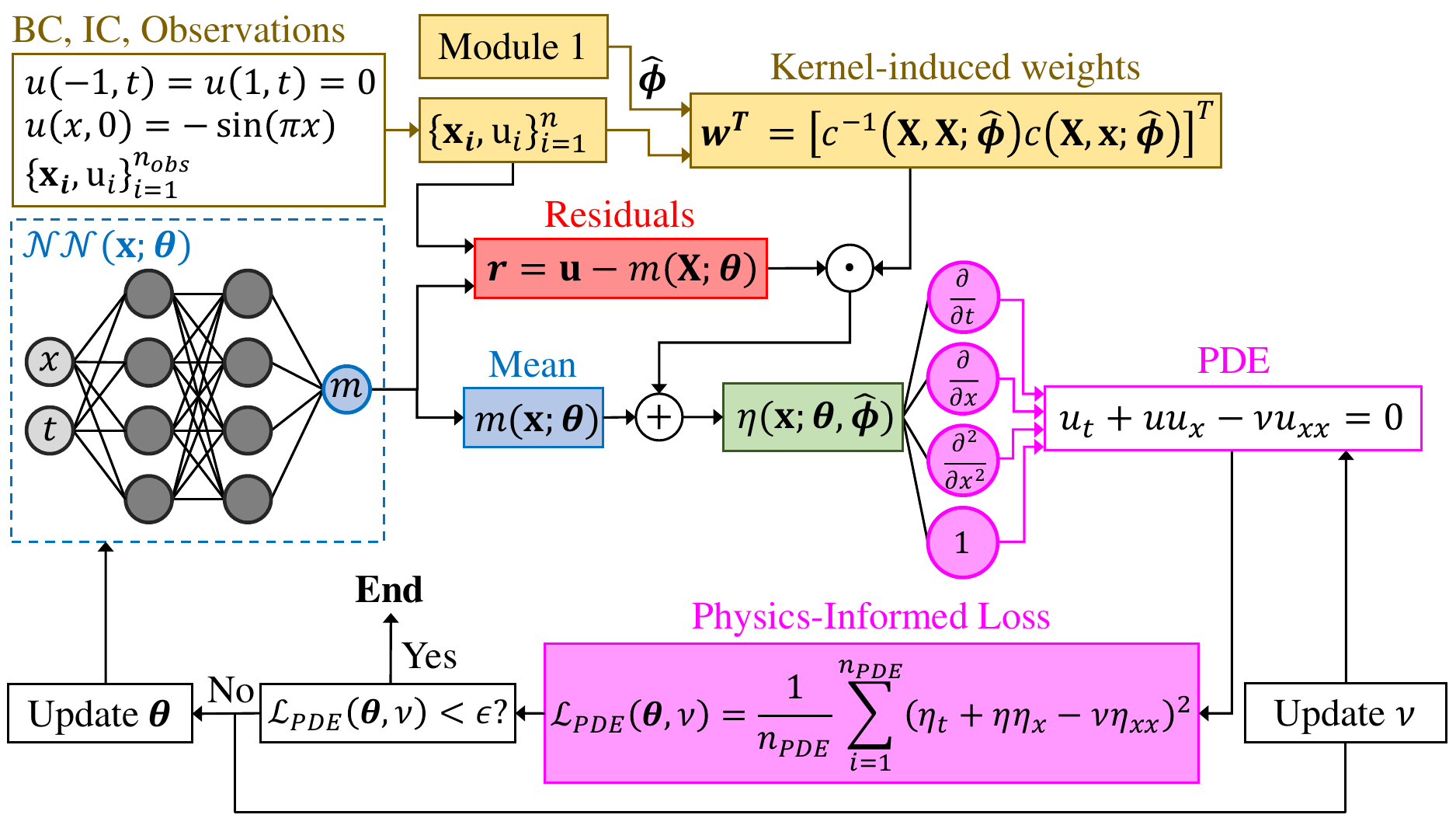}
        \captionsetup{justification=centering}
        \caption{Flowchart of module two of our framework for solving inverse problems. The flowchart is tailored to the PDE system in \Cref{eq: Burgers-main}.}
        \label{fig: flowchart-inverse}
    \end{subfigure}

    \vspace{\floatsep}  
    
    \begin{subfigure}[t]{\textwidth}
        \centering
        \includegraphics[width=1.0\textwidth]{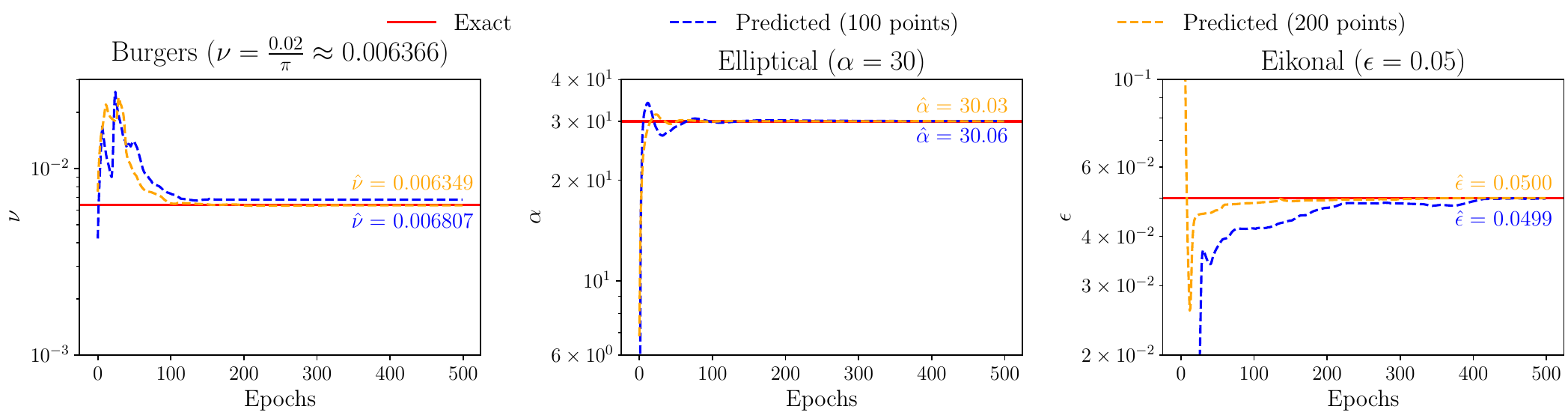}
        \captionsetup{justification=centering}
        \caption{Convergence rates are fast and improve as more data are infused into our model.}  
        \label{fig: inverse-estimation-convergence}
    \end{subfigure}
        
    \caption{\textbf{Inverse problems via NN-\short:} We modify the flowchart of module two in \Cref{fig: flowchart} in two ways to solve a PDE system whose one or more parameters may be unknown. NN-\shortspace treat observations (i.e., labeled solution data inside the domain) identically to boundary data and are very effective in using them in estimating the unknown PDE parameters.}
    \label{fig: inverse-problems}
\end{figure*}

As shown in \Cref{fig: flowchart-inverse}, we modify our framework in two ways to solve the PDE system in \Cref{eq: Burgers} assuming $\nu$ is unknown but $u(\inputb)$ is known at $n_{obs}$ random points in the domain. Specifically, we $(1)$ use the $n_{obs}$ observations in the kernel of NN-\shortspace in exactly the same way that the $n_{BC} + n_{IC}$ boundary data are handled by the kernel, and $(2)$ treat $\nu$ as one additional parameter that must be optimized along with the weights and biases of the NN.

To evaluate the performance of our approach in solving inverse problems, we consider the Burgers', Elliptic, and Eikonal PDE systems introduced in \Cref{sec: pde-description}. We solve each problem in two scenarios where there are either $n_{obs} = 100$ or $n_{obs} = 200$ observations available in the domain. 
As shown in \Cref{fig: inverse-estimation-convergence}, in all cases NN-\shortspace can estimate the unknown PDE parameter quite accurately. The convergence rate in all cases is quite fast and insignificantly reduces as $n_{obs}$ is halved from $200$ to $100$. 

    \section{Conclusions} \label{sec: conclusion}



We introduce a general framework based on Gaussian processes to solve forward and inverse problems that involve nonlinear PDEs. Our framework adds the strengths of kernels to deep NNs via kernel-weighted corrective residuals that ensure the combined model reproduces labeled data \textit{by construction}, i.e., we eliminate data loss terms from the training process of neural PDE solvers.

We design a modular, robust, and efficient framework to build NNs with kernel-weighted corrective residuals (or NN-\short \ for short) and show that the resulting model consistently outperforms competing methods on a broad range of experiments. This performance improvement is particularly impactful as our approach simplifies the training process of deep NNs while negligibly increasing the inference costs. 
As we extensively study in \Cref{sec: results}, our findings show that NN-\shortspace are not only very robust to the choice of optimizer and initial parameter values, but also applicable to a wide range of neural architectures other than FFNNs. 
We also show in \Cref{sec: inverse-problems} that NN-\shortspace can solve inverse problems with fast convergence rates. 
As proved in \Cref{sec: proof} our approach strictly satisfies the boundary and initial conditions (and reproduce the in-domain data in the case of inverse problems) as the number of sampled data tends to infinity, regardless of the geometry and the variance of the noise that may be corrupting the data.

Although we have mainly used a simple FFNN as the mean function of NN-\shortspace in our studies, we emphasize that our framework can easily accommodate any differentiable function approximator. This flexibility is demonstrated in \Cref{fig: sensitivity}, where we achieve a superior performance by leveraging a more expressive architecture as the mean function, specifically the M3 model developed in \cite{wang2021understanding}. Hence, the key takeaway is that our framework can incorporate a wide variety of parametric mean functions. One such example is HiDeNN \cite{saha2021hierarchical} which is developed for data-driven learning in the context of multiscale multi-physics problems where one can leverage hierarchical structures and decompositional learning. Another potential application is utilizing a recurrent neural network as the mean function to predict the evolution of dynamic physical systems. As demonstrated in \cite{yousefpour2024simultaneous}, our approach is also useful in addressing some of the limitations of conventional PDE solvers (e.g., the FEM) in causing local convergence in topology optimization.

The current limitation of our approach is that the contributions of the kernel-weighted \shortspace decrease in the absence of boundary data. This behavior is also observed in the LDC problem where pressure is known only at a single point on the boundary. We believe devising periodic kernels is a promising direction for addressing this limitation which will be particularly useful in multiscale simulations where PDEs with periodic BCs frequently arise in the fine-scale analyses. 

Finally, we note that our framework can be naturally extended to operator learning by reformulating its mean and covariance functions. This research direction is particularly interesting, as recent works such as \cite{batlle2024kernel} have demonstrated the power of kernel methods for learning operators in PDEs.

\section*{Acknowledgment}
We appreciate the support from the Office of Naval Research N000142312485, NASA’s Space Technology Research Grants Program 80NSSC21K1809, and National Science Foundation 2211908.
    \begin{appendices}

\setcounter{equation}{0}
\renewcommand{\theequation}{A\arabic{equation}}

\setcounter{figure}{0}
\renewcommand\thefigure{A\arabic{figure}}
\setcounter{table}{0}
\renewcommand\thetable{A\arabic{table}}

\renewcommand{\thesection}{A\arabic{section}}


\section{Properties of a Gaussian Process Surrogate} \label{sec: gp-properties}
We use an analytic one-dimensional ($1D$) function to demonstrate some of the most important characteristics of GP surrogates. Specifically, we leverage a set of examples to argue that GPs:
$(1)$ have interpretable parameters,
$(2)$ can regress or interpolate highly nonlinear functions,
$(3)$ suffer from reversion to the mean phenomena in data scarce regions,
$(4)$ can have ill-conditioned covariance matrices if their mean function interpolates the data, and
$(5)$ with manually chosen hyperparameters can faithfully surrogate a function if sufficient training samples are available. 
These properties underpin our decision for manually selecting the kernel parameters in module one of our framework. They also demonstrate the effects of a GP's mean function on its prediction power and numerical stability. 

\begin{figure}[!t]
    \centering
    \includegraphics[width=\textwidth]{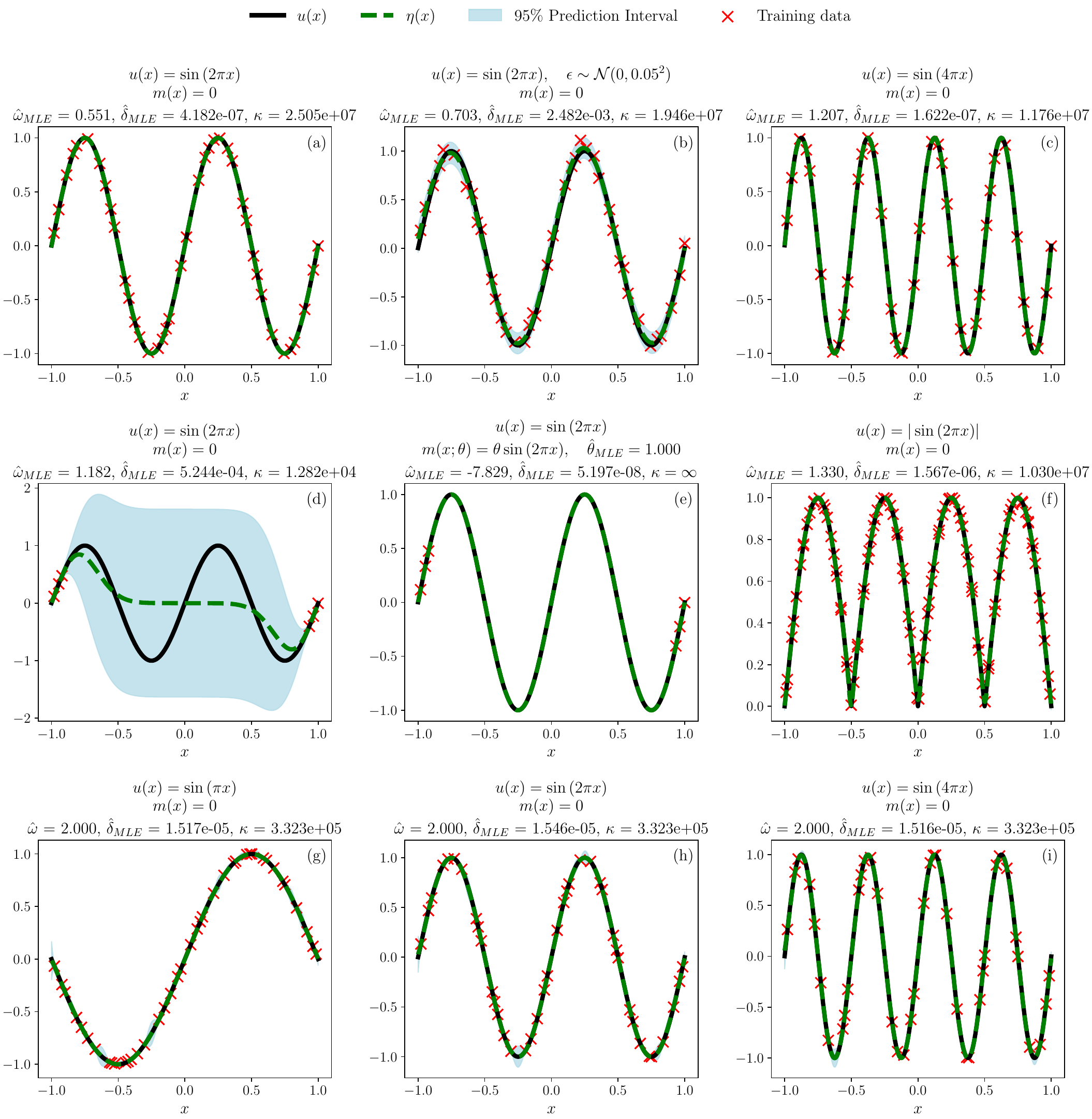}
    \caption{\textbf{Properties of Gaussian processes:} We demonstrate that GPs have interpretable hyperparameters and can regress a wide range of functions. The dependency of regression quality on the hyperparameters rapidly decreases as the size of the training data increases.}
    \label{fig: gp-plots}
\end{figure}

As demonstrated in \Cref{fig: gp-plots} our experiments involve sampling from a sinusoidal function where we study the effects of frequency, noise, data distribution, function differentiability, adopted prior mean function, and hyperparameter optimization on the behavior of GPs. For all of these studies we endow the GP with the following parametric kernel:
\begin{equation} 
    c\parens{x, x'; \sigma^2, \phi, \delta} = \sigma^2 \exp\braces{-\phi\parens{x-x'}^2} + \mathbbm{1}\{x==x'\}\delta,
    \label{eq: kernel}
\end{equation}
where $\lambdab = \brackets{\sigma^2, \phi, \delta}^T$ are the kernel parameters. In this equation, $\sigma^2$ is the process variance which, looking at Equation $1$ in the main text, does not affect the posterior mean and hence we simply set it to $1$ in our framework (this feature of our framework is in sharp contrast to other methods such as \cite{RN1886} whose performance is quite sensitive to the selected kernel parameters). 
The rest of the parameters in \Cref{eq: kernel} are defined as follows. $\phi = 10^{\omega}$ where $\omega$ is the length-scale or roughness parameter that controls the correlation strength along the $x-$axis, $\mathbbm{1}\{\cdot\}$ returns $1/0$ if the enclosed statement is true/false, and $\delta$ is the so-called nugget or jitter parameter that is added to the kernel for modeling noise and/or improving the numerical stability of the covariance matrix. We quantify the numerical stability of the covariance matrix via its condition number or $\kappa$. 

Given some training data, $\lambdab$ can be quickly estimated via maximum likelihood estimation (MLE). We denote parameter estimates obtained via this process by appending the subscript MLE to them, i.e., $\widehat\lambdab_{MLE}$. Alternatively, we can manually assign specific values to $\lambdab$. 

We first study the effect of noise by training two GPs where both GPs aim to emulate the same underlying function but one has access to noise-free responses while the other is trained on noisy data, see \Cref{fig: gp-plots} (a) and \Cref{fig: gp-plots} (b), respectively. We observe in \Cref{fig: gp-plots} (a) that the estimated value for $\widehat\delta_{MLE}$ is very small since the data is noise-free (the small value is added to reduce $\kappa$) while in \Cref{fig: gp-plots} (b) the estimated nugget parameter is much larger and close to the noise variance ($2.48\mathrm{e}{-3}$ vs. $2.50\mathrm{e}{-3}$). 
Additionally, comparing \Cref{fig: gp-plots} (a) and \Cref{fig: gp-plots} (c) we observe a direct relation between the frequency of the underlying function and the estimated kernel parameters. In particular, the magnitude of $\widehat\omega_{MLE}$ increases as $u(x)$ becomes rougher since the correlation between two points on it quickly dies out as the distance between those points increases (for this reason, $\omega$ is also sometimes called the roughness parameter). Further increasing the frequency of $u(x)$ to the extent that it resembles a noise signal directly increases $\widehat\omega_{MLE}$. 
These points indicate that the kernel parameters of a GP are interpretable.

We next study the reversion to the mean behavior and numerical instabilities of GPs in \Cref{fig: gp-plots} (d) and \Cref{fig: gp-plots} (e). In both of these scenarios the training data is only available close to the boundaries. However, we set the prior mean of the GP in \Cref{fig: gp-plots} (d) and \Cref{fig: gp-plots} (e) to zero and $m(x; \theta) = \theta \times \sin(2\pi x)$, respectively. 
The reversion to the mean behavior is clearly observed in \Cref{fig: gp-plots} (d) where the expected value of the posterior distribution is almost zero in the $(-0.5, 0.5)$ range where the correlations with the training data die out. 
The reversion to the mean behavior is also seen in \Cref{fig: gp-plots} (e) but this time it is not undesirable since the functional form of the chosen parametric mean function is similar to $u(x)$ (note that a large neural network can also reproduce the training data but such a network cannot match $u(x)$ in interior regions where there are no labeled data). This similarity forces the kernel to regress residuals that are mostly zero (i.e., the kernel must regress a constant value in the entire domain). Since any two points on a constant function have maximum correlation, regressing such residuals requires $\phi \rightarrow 0$ which, in turn, renders the covariance matrix ill-conditioned to the extent that $\kappa \rightarrow +\infty$. 
Based on these observations, in our framework we do not estimate the kernel parameters jointly with the weights and biases of the deep neural network (NN). 

Lastly, in \Cref{fig: gp-plots} (e) we demonstrate that GPs can interpolate non-differentiable functions as long as they are provided with sufficient training data. The power and efficiency of GPs in learning from data is quite robust to the hyperparameters. As shown in \Cref{fig: gp-plots} (g) through \Cref{fig: gp-plots} (i) GPs with manually selected $\omega$ can accurately surrogate $u(x)$ regardless of its frequency (the nugget value in these three cases is chosen such that $\kappa$ does not exceed a predetermined value). This attractive behavior forms the basis of our choice to manually fix $\phib$ in the first module of our framework. It is highlighted that the manual parameter selection results in sub-optimal prediction intervals but this issue does not affect our framework since we do not leverage these intervals. 

\section{Methods Description} \label{sec: methods-description}
Below, we briefly introduce the four PIML models that we have used in our comparative studies. 
To be able to directly compare the implementation of the four PIML models, we use Burgers' equation in the following descriptions. The PDE system is:
\begin{subequations}
    \begin{align}
        &\ut + u \ux - \nu \uxx = 0, && \forall x \in [-1,1], t \in (0, 1] 
        \label{eq: Burgers-pde}\\
        &u(-1, t) = u(1, t) = 0, && \forall t \in [0, 1] 
        \label{eq: Burgers-bc}\\
        &u(x, 0) = -\sin{\parens{\pi x}}, && \forall x \in [-1,1]
        \label{eq: Burgers-ic}
    \end{align}
    \label{eq: Burgers}
\end{subequations}
where $\inputb = \brackets{x, t}$ are the independent variables, $u$ is the PDE solution, and $\nu$ is a constant that denotes the kinematic viscosity.
Also, we denote the output of the NN models via $m(\inputb; \thetab)$ throughout this section. Note that we also employ $m(\inputb; \thetab)$ for denoting the NN in the mean function of NN-\short.

\subsection{Physics-informed Neural Networks (PINNs)}
\begin{figure*}[!b]
    \centering
    \begin{subfigure}[t]{0.68\textwidth}
        \centering
        \includegraphics[width=1.00\columnwidth]{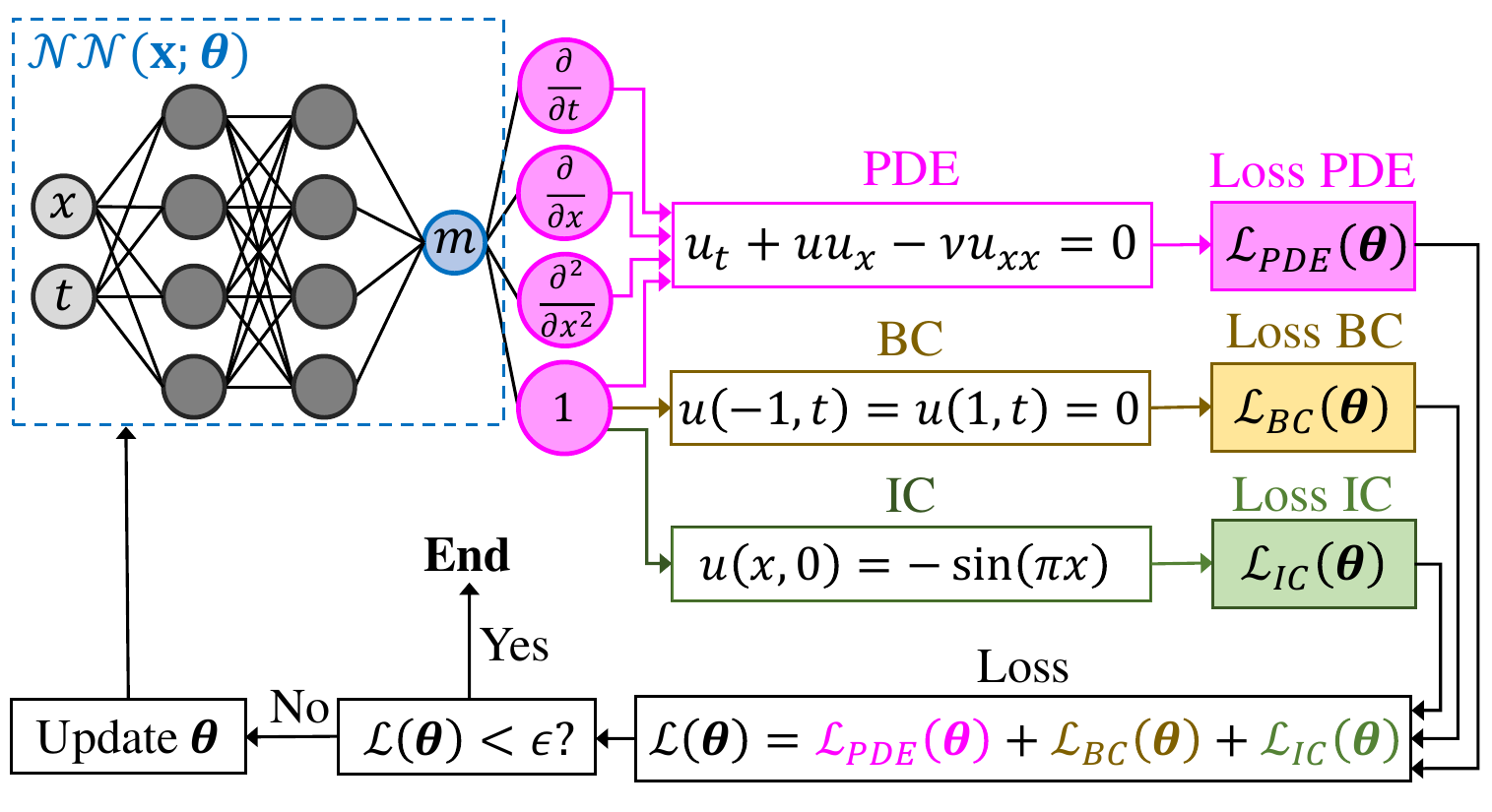}
        \captionsetup{justification=centering}
        \caption{Architecture and loss function for solving the Burgers' equation.}
        \label{fig: pinn-architecture}
    \end{subfigure}%
    \begin{subfigure}[t]{0.32\textwidth}
        \centering
        \includegraphics[width=1.00\columnwidth]{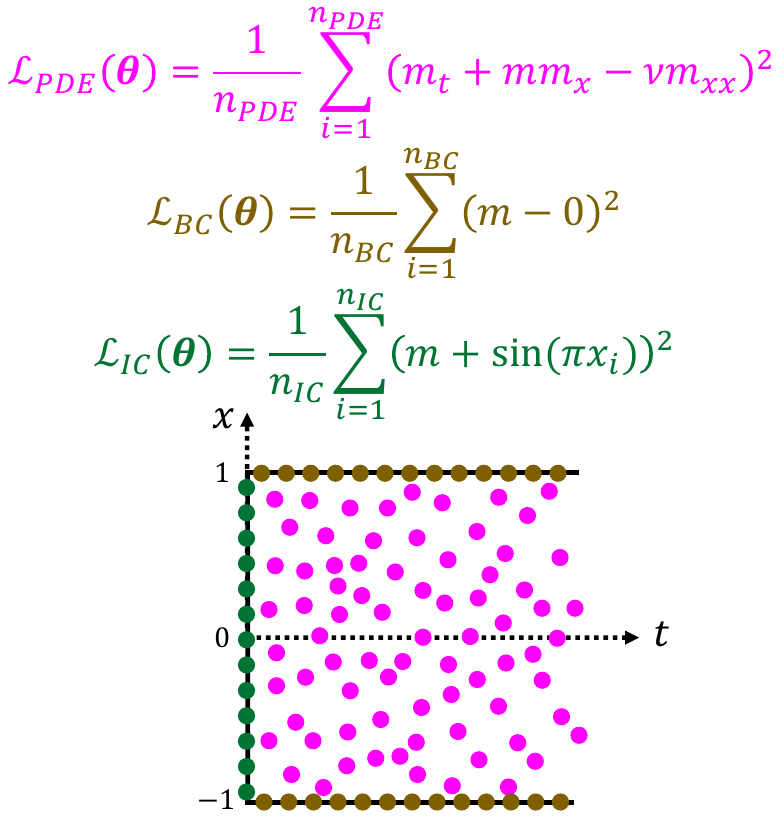}
        \captionsetup{justification=centering}
        \caption{Test points in the domain and on the boundaries.}
        \label{fig: pinn-cps}
    \end{subfigure}
    \caption{\textbf{Physics-informed neural network (PINN):} The model parameters, $\thetab$, are optimized by minimizing the three-component loss function that encourages the network to satisfy the PDE inside the domain while reproducing the initial and boundary conditions. These loss components are obtained by querying the network on a set of test points that are distributed inside the domain or on its boundaries.}
    \label{fig: pinn-flowchart}
\end{figure*}

As schematically shown in \Cref{fig: pinn-flowchart}, the essential idea of PINNs is to parameterize the relation between $u$ and $\inputb$ with a deep NN \cite{raissi2019physics}, i.e., $u(\inputb) = m(\inputb; \thetab)$ where $\thetab$ are the network's weights and biases. 
The parameters of $m$ are optimized by iteratively minimizing a loss function, denoted by $\lt$, that encourages the network to satisfy the PDE system in \Cref{eq: Burgers}. 
To calculate $\lt$, we first obtain the network's output at $n_{BC}$ points on the $x=-1$ and $x=1$ boundaries, $n_{IC}$ points on the $t=0$ boundary which marks the initial condition, and $n_{PDE}$ collocation points (CPs) inside the domain, see \Cref{fig: pinn-cps}. For the $n_{BC} + n_{IC}$ points on the boundaries, we can directly compare the network's outputs to the specified boundary and initial conditions in \Cref{eq: Burgers-bc,eq: Burgers-ic}. For each of the $n_{PDE}$ CPs, we evaluate the partial derivatives of the output and calculate the residual in \Cref{eq: Burgers-pde}. Once these three terms are calculated, we obtain $\lt$ by summing them up as follows:
\begin{equation}
    \begin{aligned}
        \lt =~ &\mathcal{L}_{PDE}(\thetab) + \mathcal{L}_{BC}(\thetab) + \mathcal{L}_{IC}(\thetab) \\
        =~ &\frac{1}{n_{PDE}}\sum_{i=1}^{n_{PDE}} \parens{m_t(\inputb_i;\thetab) + m(\inputb_i;\thetab) m_x(\inputb_i;\thetab) - \nu m_{xx}(\inputb_i;\thetab)}^2 +\\
        &\frac{1}{n_{BC}}\sum_{i=1}^{n_{BC}} \parens{m(\inputb_i;\thetab) - 0}^2 
        + \frac{1}{n_{IC}}\sum_{i=1}^{n_{IC}} \parens{m(\inputb_i;\thetab) + \sin{\parens{\pi x_i}}}^2
    \end{aligned}
    \label{eq: pinn-loss}
\end{equation}
The loss function in \Cref{eq: pinn-loss} is typically minimized via either the Adam \cite{kingma2014adam} or L-BFGS \cite{liu1989limited} methods which are both gradient-based optimization algorithms. With either Adam or L-BFGS, the parameters of the network are first initialized and then iteratively updated to minimize $\lt$. These updates rely on partial derivaties of $\lt$ with respect to $\thetab$ which can be efficiently obtained via automatic differentiation \cite{baydin2018automatic}.

While Adam and L-BFGS are both gradient-based optimization techniques, they have some major differences \cite{sun2019survey}. 
Adam is a first-order method while L-BFGS is not since it is a quasi-Newton optimization algorithm. Compared to Adam, L-BFGS is more memory-intensive and has a higher per-epoch computational cost since it uses an approximation of the Hessian matrix during the optimization. Moreover, Adam scales to large datasets better than L-BFGS which does not accommodate mini-batch training. 
However, L-BFGS typically provides lower loss values and requires fewer epochs for convergence compared to Adam. 

\subsection{Physics-informed Neural Networks With Dynamic Loss Weights}
One of the challenges associated with minimizing the loss function in \Cref{eq: pinn-loss} is that the three terms on the right-hand side disproportionately contribute to $\lt$. To mitigate this issue, a popular approach is to scale each loss component independently before summing them up, that is:
\begin{equation} 
    \lt = \mathcal{L}_{PDE}(\thetab) + w_{BC}\mathcal{L}_{BC}(\thetab) + w_{IC}\mathcal{L}_{IC}(\thetab).
    \label{eq: pinn-loss-dynamic}
\end{equation}

Since the scale of the three loss terms can change dramatically during the optimization process, these weights must be dynamic, i.e., their magnitude must be adjusted during the training. In our experiments, we follow the process described in \cite{wang2021understanding} for dynamic loss balancing and highlight that this approach is only applicable to cases where Adam is used. 

\subsection{Physics-informed Neural Networks with Hard Constraints}
An alternative approach to dynamic weight balancing is to eliminate $\mathcal{L}_{BC}(\thetab)$ and $\mathcal{L}_{IC}(\thetab)$ from \Cref{eq: pinn-loss} by requiring the model's output to satisfy the boundary and initial conditions by construction \cite{berg2018unified}. 
To this end, we now denote the output of the network by $\widetilde{m}\parens{\inputb; \thetab}$ and then formulate the final output of the model as: 
\begin{equation}
    m\parens{\inputb; \thetab} = a(\inputb)\widetilde{m}\parens{\inputb; \thetab} + b(\inputb),
    \label{eq: pinn-hc}
\end{equation}
where $a(\inputb)$ and $b(\inputb)$ are analytic functions that ensure $m\parens{\inputb; \thetab}$ satisfies \Cref{eq: Burgers-bc,eq: Burgers-ic} regardless of what $\widetilde{m}\parens{\inputb; \thetab}$ produces at $\inputb$. 
A common strategy is to choose $a(\inputb)$ to be the signed distance function that vanishes on the boundaries and produces finite values inside the domain. The construction of $b(\inputb)$ is application-specific since one has to formulate a function that satisfies the applied boundary and initial conditions while generating finite values inside the domain. For the PDE system in \Cref{eq: Burgers}, one option is $b(\inputb)=\frac{-2 \sin{\pi x}}{1+e^{-t}}$.

\subsection{Optimal Recovery}
This recent approach leverages zero-mean GPs for solving nonlinear PDEs \cite{RN1886}. Specifically, let us denote the kernel of a zero-mean GP via $c(\cdot, \cdot)$. We associate $c(\cdot, \cdot)$ with the reproducing kernel Hilbert space (RKHS) $\mathcal{U}$ where the RKHS norm is defined as $\|u\|$. Following these definitions, we can approximate $u(\inputb)$ by finding the minimizer of the following optimal recovery problem:
\begin{equation}
    \begin{aligned}
        &\underset{u \in \mathcal{U}}{\operatorname{minimize}}\ \|u\|& \\
        &\text{subject to} \\
        & \qquad \qquad \qquad  \ut(\inputb_i) + u(\inputb_i) \ux(\inputb_i) - \nu \uxx(\inputb_i) = 0, && \forall i = 1, \dots, n_{PDE} \\
        & \qquad \qquad \qquad  u(\inputb_i) = 0, && \forall i = 1, \dots, n_{BC},\\
        & \qquad \qquad \qquad u(\inputb_i) = -\sin{\parens{\pi x_i}}, && \forall i = 1, \dots, n_{IC},
    \end{aligned}    
    \label{eq: optimal-recovery}
\end{equation}
where $n_{PDE}$, $n_{BC}$, $n_{IC}$ are the number of nodes inside the domain, on the $x=-1$ and $x=1$ lines where the boundary conditions are specified, and on the $t=0$ line where the initial condition is specified, respectively. We denote the collection of these $n_{PDE} + n_{BC} + n_{IC}$ points via $\Inputb$.

The optimization problem in \Cref{eq: optimal-recovery} is infinite-dimensional and hence \cite{RN1886} leverage the representer theorem to convert it into a finite-dimensional one by defining the slack variable $\zb = \brackets{z^{(1)}, z^{(2)}, z^{(3)}, z^{(4)}}$: 
\begin{equation}
    \begin{aligned}
        &\underset{\zb \in \rspace^{N}}{\operatorname{minimize}}\ \zb^T \boldsymbol{\Theta}^{-1} \zb & \\
        &\text{subject to} \\
        & \qquad \qquad \qquad  z_i^{(2)} + z_i^{(1)}z_i^{(3)} -\nu z_i^{(4)} = 0, && \forall i = 1, \dots, n_{PDE} \\
        & \qquad \qquad \qquad  z_i^{(1)} = 0, && \forall i = 1, \dots, n_{BC},\\
        & \qquad \qquad \qquad z_i^{(1)} = -\sin{\parens{\pi x_i}}, && \forall i = 1, \dots, n_{IC},
    \end{aligned}    
    \label{eq: finite-dimensional-problem}
\end{equation}
where $N=4(n_{PDE}+n_{BC}+n_{IC})+3n_{PDE}$ and $\boldsymbol{\Theta}$ is the covariance matrix (see Section 3.4.1 of \cite{RN1886} for details on $\boldsymbol{\Theta}$). \Cref{eq: finite-dimensional-problem} can be reduced to an unconstrained optimization problem by eliminating the equality constraints following the process described in Subsection 3.3.1 of of \cite{RN1886}. Once $\zb$ is estimated, the PDE solution can be estimated at the arbitrary point $\inputb$ in the domain via GP regression.

We note that the process of defining the slack variables and obtaining the equivalent finite-dimensional optimization problem needs to be repeated for different PDE systems (e.g., in a PDE system one may have to define some of the slack variables as the Laplacian of the solution rather than the solution itself). 
Also, per the recommendations in \cite{RN1886}, $c(\cdot, \cdot)$ is  set to an anisotropic kernel and its parameters are chosen manually (i.e., they do not need to be jointly estimated with $\zb$) but, unlike our approach, this choice must be done carefully since it affects the results. In our comparative studies, we use the values reported in \cite{RN1886} for the kernel parameters. 

\section{Neural Networks with Kernel-weighted Corrective Residuals Reproduce the Data} \label{sec: proof}
\begin{proof}
We prove that the error of our model in reproducing the boundary data converges to zero as we increase the number of sampled boundary data. For the sake of completeness, we begin by a definition and invoking two theorems and then proceed with our proof. 

\begin{definition}[Reproducing kernel Hilbert space]
Let $\hsspace$ be a Hilbert space of real functions $\outputu$ defined on an index set $\inputspace$. Then, $\hsspace$ is called a Reproducing kernel Hilbert space (RKHS) with the inner product $\langle \cdot,\cdot \rangle_{\hsspace}$ if the function $c: \inputspace\times\inputspace \rightarrow \rspace$ with the following properties exists:
\begin{itemize}
    \item For any $\inputb$, $c(\inputb, \inputb')$ as a function of $\inputb'$ is in $\hsspace$,
    \item $c$ has the reproducing property, that is $\langle \outputu(\inputb'),c(\inputb', \inputb) \rangle_{\hsspace}=\outputu(\inputb)$.
\end{itemize}
Note that the norm of $\outputu$ is $\parallel \outputu \parallel_{\hsspace}=\sqrt{\langle \outputu,\outputu \rangle_{\hsspace}}$ and that $\langle c(\inputb, \cdot),c(\inputb', \cdot) \rangle_{\hsspace}=c(\inputb, \inputb')$ since both $c(\inputb', \cdot)$ and $c(\inputb, \cdot)$ are in $\hsspace$.
\end{definition}

\begin{theorem}[Mercer's Theorem]
The eigenfunctions of the real positive semidefinite kernel $c(\inputb, \inputb')$ whose eignenfunction expansion with respect to measure $\pi$ is $c(\inputb, \inputb') = \sum_{i=1}^N\alpha_i\psi_i(\inputb)\psi_i(\inputb')$, are orthonormal. That is: 
\begin{equation}
    \int \psi_i(\inputb)\psi_j(\inputb)d\pi = \delta_{ij}
    \label{eq: mercer}
\end{equation}
where $\delta_{ij}$ denotes the Kronecker delta function. 
Following this theorem, we note that for a Hilbert space defined by the linear combinations of the eigenfunctions, that is $\outputu(\inputb) = \sum_{i=1}^N \outputu_i \psi_i(\inputb)$ with $\sum_{i=1}^N {\outputu_i}/{\alpha_i}<\infty$, we have $\parallel \outputu \parallel_{\hsspace}^2=\langle \outputu,\outputu \rangle_{\hsspace}=\sum_{i=1}^N {\outputu_i}/{\alpha_i}$.
\end{theorem}

\begin{theorem}[Representer Theorem]
Each minimizer $\outputu(\inputb) \in \hsspace$ of the following functional can be represented as
$\outputu(\inputb) = \sum_{i=1}^n\alpha_i c(\inputb, \inputb_i)$:
\begin{equation}
    F[\outputu(\inputb)] = \frac{\beta}{2}\parallel \outputu(\inputb) \parallel_{\hsspace}^2 + P(\hb, \outputub).
    \label{eq: representer}
\end{equation}
where $\hb=\brackets{h_1, \cdots, h_n}^T$ is the observation vector, $\outputu(\inputb)$ denotes the function that we aim to fit to $\hb$, $\outputub=\brackets{\outputu(\inputb_1), \cdots, \outputu(\inputb_n)}^T = \brackets{\outputu_1, \cdots, \outputu_n}^T$ are the evaluations of $\outputu(\inputb)$ at configurations where $\hb$ are observed, $\beta$ is a scaling constant that balances the contributions of the two terms on the right hand side (RHS) to $F[\outputu(\inputb)]$, and $P(\cdot, \cdot)$ is a function that evaluates the quality of $\outputu(\inputb)$ in reproducing $\hb$. 
For proof of this theorem, see \cite{scholkopf2002learning, o1986automatic, kimeldorf1971some}.
\end{theorem}

In our case, to prove that our model can reproduce the boundary data we first assume that the initial and boundary conditions are sufficiently smooth functions and that the neural network (i.e., the mean function of the GP) produces finite values on the boundaries. These assumptions simplify the proof by allowing us to work with the difference of these two terms.

We now consider a specific form of \Cref{eq: representer}:
\begin{equation}
    F[\outputu(\inputb)] = \frac{1}{2}\parallel \outputu(\inputb) \parallel_{\hsspace}^2 + \frac{\lambda^2}{2} \sum_{i=1}^{n} \parens{h_i - \outputu(\inputb_i)}^2,
    \label{eq: representer2}
\end{equation}
where $\outputu(\inputb)$ is the zero-mean GP predictor and $\parallel \outputu(\inputb) \parallel_{\hsspace}$ is the RKHS norm with kernel $c(\cdot, \cdot)$. The second term on the right hand side corresponds to the negative log-likelihood of a Gaussian noise model with precision $\lambda^2$ and hence the minimizer of \Cref{eq: representer2} is the posterior mean of the GP \cite{szeliski1987regularization}. Hence, we now need to show that as $n\rightarrow\infty$ the minimizer of \Cref{eq: representer2}, which is our GP, can reproduce the data $\hb$. We denote the ground truth function that we aim to discover and the variance around it by, respectively, $h(\inputb)$ and $\tau^2(\inputb) = \int\parens{h - h(\inputb)}^2d\pi(h|\inputb)$ where $\pi(\inputb, h)$ is the probability measure that generates the data $\parens{\inputb_i, h_i}$. 

We rewrite the second term on the right hand side of \Cref{eq: representer2} as:
\begin{equation}
    \begin{split}
        &\expectation \brackets{\sum_{i=1}^{n} \parens{h_i - \outputu(\inputb_i)}^2} = 
        n \int \parens{h - \outputu(\inputb)}^2d\pi(\inputb, h) = \\
        &n \int \parens{h - h(\inputb) + h(\inputb) - \outputu(\inputb)}^2d\pi(\inputb, h) = \\
        &n \int \tau^2(\inputb)d\pi(\inputb) + 0 + n \int \parens{h(\inputb) - \outputu(\inputb)}^2d\pi(\inputb).
    \end{split}
    \label{eq: representer2-RHS}
\end{equation}
where the zero on the last line is due to the definition of $h(\inputb)$, i.e., $h(\inputb) = \expectation \brackets{h|\inputb}$. Since $\tau^2(\inputb)$ is independent of $\outputu(\inputb)$, we can use \Cref{eq: representer2-RHS} to rewrite \Cref{eq: representer2} as:
\begin{equation}
    F_{\pi}[\outputu(\inputb)] = \frac{1}{2}\parallel \outputu(\inputb) \parallel_{\hsspace}^2 + \frac{n\lambda^2}{2} \int \parens{h(\inputb) - \outputu(\inputb)}^2d\pi(\inputb).
    \label{eq: representer3}
\end{equation}

We now invoke Mercer's theorem to write $\outputu(\inputb) = \sum_{i=1}^{\infty} \outputu_i \psi_i(\inputb)$ and $h(\inputb) = \sum_{i=1}^{\infty} h_i \psi_i(\inputb)$ where $\psi_i$ are the eigenfunctions of the nondegenerage kernel of the GP. Since $\braces{\psi_i}$ form an orthonormal basis, we can write:
\begin{equation}
    F_{\pi}[\outputu(\inputb)] = \frac{1}{2} \sum_{i=1}^{\infty} \frac{\outputu_i^2}{\alpha_i}
    + \frac{n\lambda^2}{2} \sum_{i=1}^{\infty} \parens{h_i - \outputu_i}^2.
    \label{eq: representer4}
\end{equation}
We take the derivative of \Cref{eq: representer4} with respect to $\outputu_i$ and set it to zero to obtain:
\begin{equation}
    \outputu_i = \frac{\alpha_i h_i}{\alpha_i + {1}/{n\lambda^2}}.
    \label{eq: g-limit}
\end{equation}
Since ${1}/{n\lambda^2} \rightarrow 0$ as $n\rightarrow\infty$, in the limit $\outputu_i \rightarrow h_i$, i.e., our zero-mean GP predictor corrects for the error that $m(\inputb, \thetab)$ has on reproducing the initial and boundary conditions. Note that the convergence in \Cref{eq: g-limit} does not depend on $\tau^2(\inputb)$ and hence holds for the case where the observation vector $\hb$ is noisy. 
\end{proof}

\section{Additional Experiments}\label{sec: additionalResults}
In the following subsections, we summarize the findings from additional experiments in forward problems. 
First, we provide further details on the results obtained for the LDC problem. Next, we present the results obtained using NN-\shortspace for the 2D Helmholtz equation and the inviscid Burgers' equation. The latter is a hyperbolic PDE, and hence broadens the range of PDEs addressed in \Cref{sec: results}. Finally, we compare and discuss the computational cost of our method with that of PINN.

\subsection{Lid-driven Cavity Problem}
The solution of the LDC problem consists of three dependent variables which are the pressure $p(\inputb)$ and the two velocity components in the $x$ and $y$ directions, $u(\inputb)$ and $v(\inputb)$, respectively. In the main text we report the mean of the Euclidean norm of the error on the three outputs (see \Cref{tab: comparison-main} in the main text). In \Cref{tab: comparison-ldc} we provide the errors for the individual outputs of this benchmark problem and observe the same trend where NN-\shortspace consistently outperforms other methods. 
We also notice that all the models predict pressure with less accuracy compared to the velocity components. This trend is due to the facts that not only the scale of $p(x, y)$ is smaller than the velocity components, but also $p(x, y)$ is known at a single point on the boundaries whereas $u(x, y)$ and $v(x, y)$ are known everywhere on the boundaries. 
\begin{table}[!t]
    \tiny
    \caption{\textbf{Summary of comparative studies for the LDC problem:} We report $L_{2,e}$ of different methods as a function of model capacity and $A$. The symbol $\otimes$ indicates the network architecture (e.g., $4\otimes 10$ is an NN which has four $10-$ neuron hidden layers). Unlike NN-based methods, \GPor's accuracy relies on the number of interior nodes which we set to $1{,}000$ or $2{,}000$. \GPorSpace is not applied to LDC as it relies on manual derivation of the equivalent variational problem which, unlike the first three PDEs, is not done by the developers \cite{RN1886}. }    
    \label{tab: comparison-ldc}
    \begin{tabular*}{\textwidth}{@{\extracolsep\fill}lccccccccccc}
        \toprule%
        & \multicolumn{2}{@{}c@{}}{\textbf{NN-\short}} & \multicolumn{2}{@{}c@{}}{\textbf{GP}$_\text{\textbf{OR}}$}  & \multicolumn{2}{@{}c@{}}{\textbf{PINN}} & \multicolumn{2}{@{}c@{}}{\textbf{PINN}$_\text{\textbf{DL}}$} & \multicolumn{2}{@{}c@{}}{\textbf{PINN}$_\text{\textbf{HC}}$} \\
        \cmidrule{2-3}\cmidrule{4-5} \cmidrule{6-7} \cmidrule{8-9} \cmidrule{10-11}%
        \diagbox[width=10em]{\textbf{Problem}}{\textbf{Capacity}} & $4\otimes 10$ & $4\otimes 20$ & $1{,}000$ & $2{,}000$ & $4\otimes 10$ & $4\otimes 20$ & $4\otimes 10$ & $4\otimes 20$ & $4\otimes 10$ & $4\otimes 20$ \\
        \midrule
        LDC ($A=3$) \quad \Centerstack{$u$ \\$v$ \\ $p$}  & \Centerstack{$1.92\mathrm{e}{-1}$\\ $1.74\mathrm{e}{-1}$ \\ $1.91\mathrm{e}{-1}$} & \Centerstack{$\mathbf{8.56\mathbf{e}{-2}}$\\ $\mathbf{8.25\mathbf{e}{-2}}$ \\ $\mathbf{9.19\mathbf{e}{-2}}$} & \Centerstack{$-$\\ $-$ \\ $-$} & \Centerstack{$-$\\ $-$ \\ $-$} & \Centerstack{$2.66\mathrm{e}{-1}$\\ $2.78\mathrm{e}{-1}$ \\ $2.72\mathrm{e}{-1}$} & \Centerstack{$1.23\mathrm{e}{-1}$\\ $1.28\mathrm{e}{-1}$ \\ $1.33\mathrm{e}{-1}$} & \Centerstack{$2.99\mathrm{e}{-1}$\\ $3.07\mathrm{e}{-1}$ \\ $2.98\mathrm{e}{-1}$} & \Centerstack{$1.26\mathrm{e}{-1}$\\ $1.23\mathrm{e}{-1}$ \\ $1.25\mathrm{e}{-1}$} & \Centerstack{$3.99\mathrm{e}{-1}$\\ $3.06\mathrm{e}{-1}$ \\ $5.92\mathrm{e}{-1}$} & \Centerstack{$4.97\mathrm{e}{-1}$\\ $3.95\mathrm{e}{-1}$ \\ $6.96\mathrm{e}{-1}$} \\ \hline
        LDC ($A=5$) \quad \Centerstack{$u$\\$v$ \\ $p$}  & \Centerstack{$2.49\mathrm{e}{-1}$\\ $2.51\mathrm{e}{-1}$ \\ $4.33\mathrm{e}{-1}$} & \Centerstack{$\mathbf{2.22\mathbf{e}{-1}}$\\ $\mathbf{2.20\mathbf{e}{-1}}$ \\ $\mathbf{3.94\mathbf{e}{-1}}$} & \Centerstack{$-$\\ $-$ \\ $-$} & \Centerstack{$-$\\ $-$ \\ $-$} & \Centerstack{$6.01\mathrm{e}{-1}$\\ $6.32\mathrm{e}{-1}$ \\ $9.17\mathrm{e}{-1}$} & \Centerstack{$5.67\mathrm{e}{-1}$\\ $5.92\mathrm{e}{-1}$ \\ $8.72\mathrm{e}{-1}$} & \Centerstack{$5.97\mathrm{e}{-1}$\\ $6.29\mathrm{e}{-1}$ \\ $9.23\mathrm{e}{-1}$} & \Centerstack{$5.18\mathrm{e}{-1}$\\ $5.43\mathrm{e}{-1}$ \\ $8.09\mathrm{e}{-1}$} & \Centerstack{$1.02\mathrm{e}{0}$\\ $7.04\mathrm{e}{-1}$ \\ $1.36\mathrm{e}{0}$} & \Centerstack{$7.57\mathrm{e}{-1}$\\ $5.64\mathrm{e}{-1}$ \\ $1.41\mathrm{e}{0}$} \\ \bottomrule
    \end{tabular*}
\end{table}

\subsection{2D Helmholtz Equation}
\begin{figure*}[!b]    
    \centering
    \includegraphics[width=0.8\textwidth]{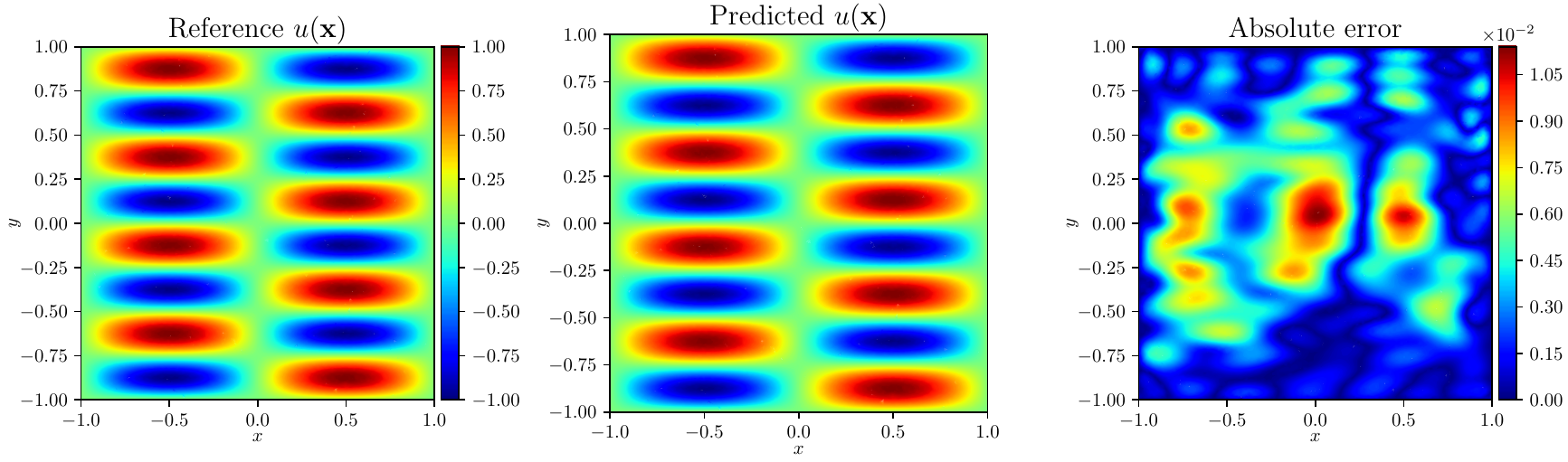}
    \caption{\textbf{Reference solutions, predicted solution and error map on the Helmholtz equation:} Our approach with a $4 \otimes 20$ architecture provides much lower errors compared to other methods and automatically adapts to high- and low-frequency solutions.}
    \label{fig: helmholtz}
\end{figure*}

In \Cref{fig: helmholtz} we solve a canonical PDE system known as Helmholtz \cite{wang2021understanding} which is defined as:
\begin{equation}
    \begin{aligned}
        & \uxx(x, y) + \uyy(x, y) + u(x, y)= q(x, y), && \forall x,y \in (-1,1)^2\\
        &u(x, -1) = u(x, 1) = 0, && \forall x \in [-1, 1] \\
        &u(-1, y) = u(1, y) = 0, && \forall y \in [-1, 1]
    \end{aligned}
    \label{eq: helmholtz}
\end{equation}

In \Cref{eq: helmholtz}, $q(x, y)$ is constructed such that the analytic solution is $u(x, y) = \sin(a_1\pi x) \sin(a_2 \pi y)$ where $a_1$ and $a_2$ are two constants that control the frequency along the $x$ and $y$ directions, respectively. The Helmholtz equation is a well-studied benchmark problem since PINNs fail to accurately solve it. To address this shortcoming, recent works have introduced quite complex architectures which typically leverage adaptive loss functions. We test our framework on this benchmark problem by setting $a_1=1$ and $a_2=4$ while using the same architecture and training procedure that are used in our comparative studies. As shown in \Cref{fig: helmholtz} our predictions accurately capture both the high- and low-frequency features of the solution.
We note that the solution in \Cref{fig: helmholtz} is 5 times more accurate than the one reported in \cite{wang2021understanding} which employs a considerably larger architecture ($4 \otimes 50$) and leverages the adaptive loss function described in \Cref{eq: pinn-loss-dynamic}.
\subsection{Inviscid Burgers' equation} \label{sec inviscid burgers}
The PDEs we addressed in \Cref{sec: results} are either parabolic or elliptic. To demonstrate that our method is also effective for hyperbolic PDEs, we apply it to the inviscid Burgers' equation, \ie \Cref{eq: Burgers-main} with $\nu=0$. For this purpose, we incorporate the training strategy proposed in \cite{liu2024discontinuity} into our GP framework. Specifically, the authors suggest using the following loss function for this problem:
\begin{equation} 
    \lt = w_{PDE}\mathcal{L}_{PDE}(\thetab) + w_{IBCs}\mathcal{L}_{IBCs}(\thetab) + w_{RH}\mathcal{L}_{RH}(\thetab)
    \label{eq: loss-inviscid-burgers}
\end{equation}
where $\mathcal{L}_{RH}(\thetab)$ is a novel term based on the Rankine-Hugoniot relation constraint, which can be computed as:
\begin{equation}
    \mathcal{L}_{RH}(\thetab) = \frac{1}{n_{RH}}\sum_{i=1}^{n_{RH}}(\eta(x=0,t)-\eta(x=0,t=0))^2.
    \label{eq l-rh loss}
\end{equation}

The authors of \cite{liu2024discontinuity} also propose to scale the contribution of each collocation point in $\mathcal{L}_{PDE}(\thetab)$ based on their gradients so that points in smooth regions are prioritized during training:
\begin{equation}
    \mathcal{L}_{PDE}(\thetab) = \frac{1}{n_{PDE}} \sum_{i=1}^{n_{PDE}} \parens{\lambda_i(\eta_{t_i} + \eta_i \eta_{x_i})}^2
\end{equation}
where 
\begin{equation}
    \lambda_i = \frac{1}{k_1 \parens{\bars{\eta_{x_i}}-\eta_{x_i}}+1}.
\end{equation}
The authors suggest the range $0.1 \leq k_1 \leq 0.4$, so we randomly selected $k_1 = 0.2$ for our studies. We note that we can remove the term $\mathcal{L}_{IBCs}(\thetab)$ in the loss in \Cref{eq: loss-inviscid-burgers} since our approach automatically satisfies the IC and BCs thanks to the kernels. Therefore, we just need to add $\mathcal{L}_{RH}(\thetab)$ to our original loss function and weight the residuals of collocation points as described above.

\begin{figure*}[!t]
    \centering
    \begin{subfigure}[t]{\textwidth}
        \centering
        \includegraphics[width=1.0\textwidth]{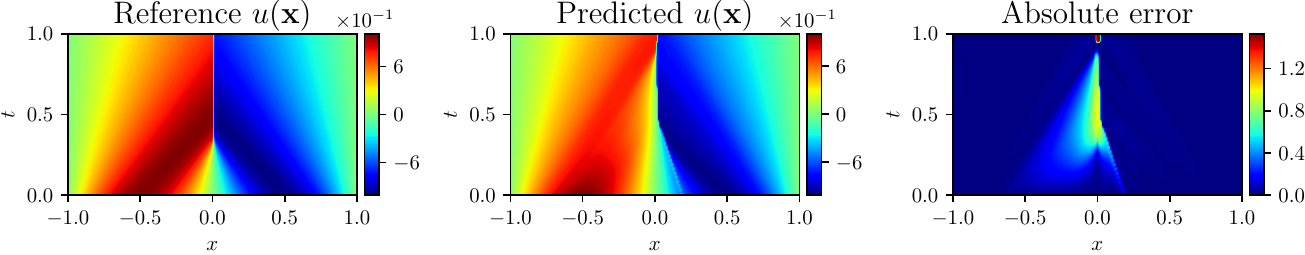}
        \captionsetup{justification=centering}
        \caption{We solve the inviscid Burgers' equation by integrating a naive training strategy within NN-\short, resulting in a test error  $L_{2,e} =  1.38\mathrm{e}{-1}$.}
        \label{fig inviscid Burgers Naive}
    \end{subfigure}
    \vspace{\floatsep}
    \begin{subfigure}[t]{\textwidth}
        \centering
        \includegraphics[width=1.0\textwidth]{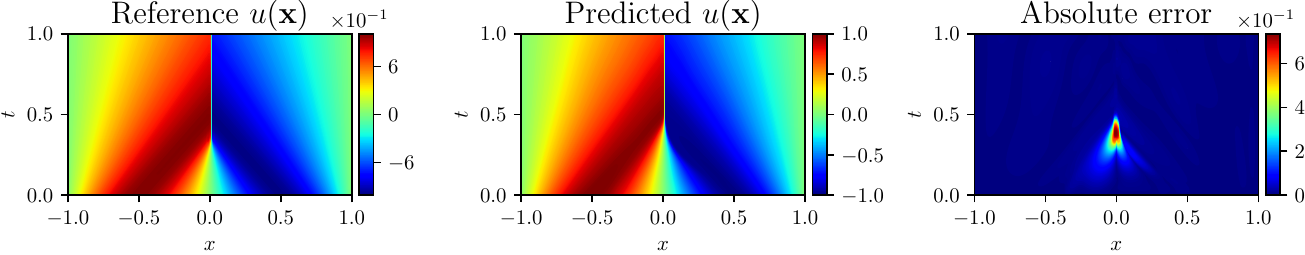}
        \captionsetup{justification=centering}
        \caption{We solve the inviscid Burgers' equation by integrating the proposed training strategy in \cite{liu2024discontinuity} within NN-\short, resulting in a test error $L_{2,e} = 3.65\mathrm{e}{-2}$.} 
        \label{fig inviscid Burgers Novel}
    \end{subfigure}
    
    \caption{\textbf{1D inviscid Burgers' equation:} We compare the performance of a naive loss function with a tailored loss function specifically designed for this problem, as proposed in \cite{liu2024discontinuity}. By incorporating this novel training mechanism into our framework, we achieve an order of magnitude lower error compared to the naive training approach.}
    \label{fig inviscid Burgers}
\end{figure*}

The results of integrating the approach of \cite{liu2024discontinuity} within NN-CoRes are shown in \Cref{fig inviscid Burgers}, where the reference solution is obtained via the Lax-Wendroff scheme \cite{lax2005systems}.
In this figure, we show two scenarios: in \Cref{fig inviscid Burgers Naive} we use a vanilla PINN as the mean function in our approach while in \Cref{fig inviscid Burgers Novel} we use the approach that was described above. 
Our results show that a suitable training mechanism for solving a hyperbolic PDE such as the inviscid Burgers' equation can be easily incorporated within our GP-based framework to obtain a superior performance to that of a naive approach. One more time, the studies carried out in this section show a major contribution of our work, which is that any researcher can integrate their developments within our GP-based framework. These studies also indicate that some of the fundamental limitations of PIML persist in our GP-based framework too, i.e., depending on the PDE system our loss function may also need additional PDE-dependent terms. For example, for solving the inviscid Burgers' equation with a non-stationary shock, we can no longer augment the loss function via \Cref{eq l-rh loss} which is specifically developed for a stationary shock at $x=0$ (i.e., a new loss term would be needed).

\subsection{Computational Cost} 
\label{sec: comp-cost}

To provide a benchmark for the computational cost of NN-\short, we report in \Cref{table: time-comparison} its training time per epoch compared to PINN (using the same architecture and optimizer settings). As it can be observed, the time per epoch of NN-\shortspace is slightly higher compared to PINN across different problems. We attribute this behavior to the fact that our method needs to evaluate the kernels in addition to the NN-based mean function for making predictions. While this slightly increases the inference costs, it offers the advantage of automatically satisfying the BCs/IC and facilitating the training of the NN by only requiring one term in the loss function (see \Cref{fig: flowchart}). This leads to faster convergence and higher accuracies across all problems, as demonstrated in \Cref{fig: loss-convergence-error-decomposition} and \ref{fig: loss-convergence-error-decomposition-ldc}.

\begin{table}[!b]
\small
    \caption{\textbf{Training time in seconds per epoch for NN-\shortspace and PINN across different problems with L-BFGS:} NN-\shortspace takes slightly more time per epoch across all problems compared to the baseline PINN. However, the curves shown in \Cref{fig: loss-convergence-error-decomposition} and \ref{fig: loss-convergence-error-decomposition-ldc} show that NN-\shortspace requires fewer epochs to converge than PINN while consistently achieving higher accuracies. The symbol $\otimes$ indicates the network architecture (e.g., $4\otimes 20$ is an NN which has four $20-$ neuron hidden layers).}
    \label{tab:comparison-results}
    \begin{tabular*}{\textwidth}{@{\extracolsep\fill}lccccc}
        \toprule
        \diagbox[width=10em]{\textbf{Model}}{\textbf{Problem}} & Burgers' ($\nu = \frac{0.01}{\pi}$) & Elliptic ($\alpha = 30$) & Eikonal ($\epsilon = 0.01$) & LDC ($A=5$) \\
        \midrule
        NN-\shortspace ($4 \otimes 20$) & $0.91$ & $1.18$ & $1.11$ & $1.94$  \\
        PINN ($4 \otimes 20$) & $0.70$ & $0.84$ & $0.83$ & $1.24$  \\
        \bottomrule
    \end{tabular*}
\label{table: time-comparison}
\end{table}

\end{appendices}
    \pagebreak
    \bibliographystyle{unsrt} 
    \bibliography{00_Ref.bib}     

\begin{thebibliography}{10}

\bibitem{RN995}
S.~L. Brunton, J.~L. Proctor, and J.~N. Kutz.
\newblock Discovering governing equations from data by sparse identification of nonlinear dynamical systems.
\newblock {\em Proc Natl Acad Sci U S A}, 113(15):3932--7, 2016.

\bibitem{schaeffer2013sparse}
Hayden Schaeffer, Russel Caflisch, Cory~D Hauck, and Stanley Osher.
\newblock Sparse dynamics for partial differential equations.
\newblock {\em Proceedings of the National Academy of Sciences}, 110(17):6634--6639, 2013.

\bibitem{RN1024}
M.~Mozaffar, R.~Bostanabad, W.~Chen, K.~Ehmann, J.~Cao, and M.~A. Bessa.
\newblock Deep learning predicts path-dependent plasticity.
\newblock {\em Proc Natl Acad Sci U S A}, 116(52):26414--26420, 2019.

\bibitem{RN829}
S.~Rahimi-Aghdam, V.~T. Chau, H.~Lee, H.~Nguyen, W.~Li, S.~Karra, E.~Rougier, H.~Viswanathan, G.~Srinivasan, and Z.~P. Bazant.
\newblock Branching of hydraulic cracks enabling permeability of gas or oil shale with closed natural fractures.
\newblock {\em Proc Natl Acad Sci U S A}, 116(5):1532--1537, 2019.

\bibitem{RN936}
Y.~Bar-Sinai, S.~Hoyer, J.~Hickey, and M.~P. Brenner.
\newblock Learning data-driven discretizations for partial differential equations.
\newblock {\em Proc Natl Acad Sci U S A}, 116(31):15344--15349, 2019.

\bibitem{RN1311}
S.~Rasp, M.~S. Pritchard, and P.~Gentine.
\newblock Deep learning to represent subgrid processes in climate models.
\newblock {\em Proc Natl Acad Sci U S A}, 115(39):9684--9689, 2018.

\bibitem{santolini2018predicting}
Marc Santolini and Albert-L{\'a}szl{\'o} Barab{\'a}si.
\newblock Predicting perturbation patterns from the topology of biological networks.
\newblock {\em Proceedings of the National Academy of Sciences}, 115(27):E6375--E6383, 2018.

\bibitem{lucor2022simple}
Didier Lucor, Atul Agrawal, and Anne Sergent.
\newblock Simple computational strategies for more effective physics-informed neural networks modeling of turbulent natural convection.
\newblock {\em Journal of Computational Physics}, 456:111022, 2022.

\bibitem{fang2023physics}
Qian Fang, Xuankang Mou, and Shiben Li.
\newblock A physics-informed neural network based on mixed data sampling for solving modified diffusion equations.
\newblock {\em Scientific Reports}, 13(1):2491, 2023.

\bibitem{jagtap2022physics}
Ameya~D Jagtap, Zhiping Mao, Nikolaus Adams, and George~Em Karniadakis.
\newblock Physics-informed neural networks for inverse problems in supersonic flows.
\newblock {\em Journal of Computational Physics}, 466:111402, 2022.

\bibitem{pun2019physically}
GP~Purja Pun, R~Batra, R~Ramprasad, and Y~Mishin.
\newblock Physically informed artificial neural networks for atomistic modeling of materials.
\newblock {\em Nature communications}, 10(1):2339, 2019.

\bibitem{lotfollahi2023biologically}
Mohammad Lotfollahi, Sergei Rybakov, Karin Hrovatin, Soroor Hediyeh-Zadeh, Carlos Talavera-L{\'o}pez, Alexander~V Misharin, and Fabian~J Theis.
\newblock Biologically informed deep learning to query gene programs in single-cell atlases.
\newblock {\em Nature Cell Biology}, 25(2):337--350, 2023.

\bibitem{RN1958}
Raphaël Pestourie, Youssef Mroueh, Chris Rackauckas, Payel Das, and Steven~G. Johnson.
\newblock Physics-enhanced deep surrogates for partial differential equations.
\newblock {\em Nature Machine Intelligence}, 5(12):1458--1465, 2023.

\bibitem{kozuch2018combined}
Daniel~J Kozuch, Frank~H Stillinger, and Pablo~G Debenedetti.
\newblock Combined molecular dynamics and neural network method for predicting protein antifreeze activity.
\newblock {\em Proceedings of the National Academy of Sciences}, 115(52):13252--13257, 2018.

\bibitem{coin2003enhanced}
Lachlan Coin, Alex Bateman, and Richard Durbin.
\newblock Enhanced protein domain discovery by using language modeling techniques from speech recognition.
\newblock {\em Proceedings of the National Academy of Sciences}, 100(8):4516--4520, 2003.

\bibitem{curtarolo2013high}
Stefano Curtarolo, Gus~LW Hart, Marco~Buongiorno Nardelli, Natalio Mingo, Stefano Sanvito, and Ohad Levy.
\newblock The high-throughput highway to computational materials design.
\newblock {\em Nature materials}, 12(3):191--201, 2013.

\bibitem{butler2018machine}
Keith~T Butler, Daniel~W Davies, Hugh Cartwright, Olexandr Isayev, and Aron Walsh.
\newblock Machine learning for molecular and materials science.
\newblock {\em Nature}, 559(7715):547--555, 2018.

\bibitem{hart2021machine}
Gus~LW Hart, Tim Mueller, Cormac Toher, and Stefano Curtarolo.
\newblock Machine learning for alloys.
\newblock {\em Nature Reviews Materials}, 6(8):730--755, 2021.

\bibitem{shi2019deep}
Zhe Shi, Evgenii Tsymbalov, Ming Dao, Subra Suresh, Alexander Shapeev, and Ju~Li.
\newblock Deep elastic strain engineering of bandgap through machine learning.
\newblock {\em Proceedings of the National Academy of Sciences}, 116(10):4117--4122, 2019.

\bibitem{RN666}
W.~K. Lee, S.~Yu, C.~J. Engel, T.~Reese, D.~Rhee, W.~Chen, and T.~W. Odom.
\newblock Concurrent design of quasi-random photonic nanostructures.
\newblock {\em Proc Natl Acad Sci U S A}, 114(33):8734--8739, 2017.

\bibitem{liu2023special}
Wing~Kam Liu, Miguel~A Bessa, Francisco Chinesta, Shaofan Li, and Nathaniel Trask.
\newblock Special issue of computational mechanics on machine learning theories, modeling, and applications to computational materials science, additive manufacturing, mechanics of materials, design and optimization.
\newblock {\em Computational Mechanics}, 72(1):1--2, 2023.

\bibitem{gebru2017using}
Timnit Gebru, Jonathan Krause, Yilun Wang, Duyun Chen, Jia Deng, Erez~Lieberman Aiden, and Li~Fei-Fei.
\newblock Using deep learning and google street view to estimate the demographic makeup of neighborhoods across the united states.
\newblock {\em Proceedings of the National Academy of Sciences}, 114(50):13108--13113, 2017.

\bibitem{thirunavukarasu2023large}
Arun~James Thirunavukarasu, Darren Shu~Jeng Ting, Kabilan Elangovan, Laura Gutierrez, Ting~Fang Tan, and Daniel Shu~Wei Ting.
\newblock Large language models in medicine.
\newblock {\em Nature medicine}, 29(8):1930--1940, 2023.

\bibitem{RN1116}
L.~Lu, M.~Dao, P.~Kumar, U.~Ramamurty, G.~E. Karniadakis, and S.~Suresh.
\newblock Extraction of mechanical properties of materials through deep learning from instrumented indentation.
\newblock {\em Proc Natl Acad Sci U S A}, 117(13):7052--7062, 2020.

\bibitem{RN1637}
Hengjie Wang, Robert Planas, Aparna Chandramowlishwaran, and Ramin Bostanabad.
\newblock Mosaic flows: A transferable deep learning framework for solving pdes on unseen domains.
\newblock {\em Computer Methods in Applied Mechanics and Engineering}, 389:114424, 2022.

\bibitem{RN1929}
Ziad Aldirany, Régis Cottereau, Marc Laforest, and Serge Prudhomme.
\newblock Multi-level neural networks for accurate solutions of boundary-value problems.
\newblock {\em Computer Methods in Applied Mechanics and Engineering}, 419:116666, 2024.

\bibitem{von2022mean}
Jakob~GR von Saldern, Johann~Moritz Reumsch{\"u}ssel, Thomas~L Kaiser, Moritz Sieber, and Kilian Oberleithner.
\newblock Mean flow data assimilation based on physics-informed neural networks.
\newblock {\em Physics of Fluids}, 34(11), 2022.

\bibitem{RN1835}
George~Em Karniadakis, Ioannis~G. Kevrekidis, Lu~Lu, Paris Perdikaris, Sifan Wang, and Liu Yang.
\newblock Physics-informed machine learning.
\newblock {\em Nature Reviews Physics}, 3(6):422--440, 2021.

\bibitem{djeridane2006neural}
Badis Djeridane and John Lygeros.
\newblock Neural approximation of pde solutions: An application to reachability computations.
\newblock In {\em Proceedings of the 45th IEEE Conference on Decision and Control}, pages 3034--3039. IEEE, 2006.

\bibitem{lagaris1998artificial}
Isaac~E Lagaris, Aristidis Likas, and Dimitrios~I Fotiadis.
\newblock Artificial neural networks for solving ordinary and partial differential equations.
\newblock {\em IEEE transactions on neural networks}, 9(5):987--1000, 1998.

\bibitem{raissi2019physics}
Maziar Raissi, Paris Perdikaris, and George~E Karniadakis.
\newblock Physics-informed neural networks: A deep learning framework for solving forward and inverse problems involving nonlinear partial differential equations.
\newblock {\em Journal of Computational physics}, 378:686--707, 2019.

\bibitem{RN1926}
Justin Sirignano and Konstantinos Spiliopoulos.
\newblock Dgm: A deep learning algorithm for solving partial differential equations.
\newblock {\em Journal of computational physics}, 375:1339--1364, 2018.

\bibitem{mcclenny2020self}
Levi McClenny and Ulisses Braga-Neto.
\newblock Self-adaptive physics-informed neural networks using a soft attention mechanism.
\newblock {\em arXiv preprint arXiv:2009.04544}, 2020.

\bibitem{jagtap2020adaptive}
Ameya~D Jagtap, Kenji Kawaguchi, and George~Em Karniadakis.
\newblock Adaptive activation functions accelerate convergence in deep and physics-informed neural networks.
\newblock {\em Journal of Computational Physics}, 404:109136, 2020.

\bibitem{wang2021understanding}
Sifan Wang, Yujun Teng, and Paris Perdikaris.
\newblock Understanding and mitigating gradient flow pathologies in physics-informed neural networks.
\newblock {\em SIAM Journal on Scientific Computing}, 43(5):A3055--A3081, 2021.

\bibitem{bu2021quadratic}
Jie Bu and Anuj Karpatne.
\newblock Quadratic residual networks: A new class of neural networks for solving forward and inverse problems in physics involving pdes.
\newblock In {\em Proceedings of the 2021 SIAM International Conference on Data Mining (SDM)}, pages 675--683. SIAM, 2021.

\bibitem{gao2021phygeonet}
Han Gao, Luning Sun, and Jian-Xun Wang.
\newblock Phygeonet: Physics-informed geometry-adaptive convolutional neural networks for solving parameterized steady-state pdes on irregular domain.
\newblock {\em Journal of Computational Physics}, 428:110079, 2021.

\bibitem{jagtap2021extended}
Ameya~D Jagtap and George~E Karniadakis.
\newblock Extended physics-informed neural networks (xpinns): A generalized space-time domain decomposition based deep learning framework for nonlinear partial differential equations.
\newblock In {\em AAAI spring symposium: MLPS}, volume~10, 2021.

\bibitem{baydin2018automatic}
Atilim~Gunes Baydin, Barak~A Pearlmutter, Alexey~Andreyevich Radul, and Jeffrey~Mark Siskind.
\newblock Automatic differentiation in machine learning: a survey.
\newblock {\em Journal of Marchine Learning Research}, 18:1--43, 2018.

\bibitem{chen2018gradnorm}
Zhao Chen, Vijay Badrinarayanan, Chen-Yu Lee, and Andrew Rabinovich.
\newblock Gradnorm: Gradient normalization for adaptive loss balancing in deep multitask networks.
\newblock In {\em International conference on machine learning}, pages 794--803. PMLR, 2018.

\bibitem{van2022optimally}
Remco van~der Meer, Cornelis~W Oosterlee, and Anastasia Borovykh.
\newblock Optimally weighted loss functions for solving pdes with neural networks.
\newblock {\em Journal of Computational and Applied Mathematics}, 405:113887, 2022.

\bibitem{RN1530}
Sifan Wang, Yujun Teng, and Paris Perdikaris.
\newblock Understanding and mitigating gradient flow pathologies in physics-informed neural networks.
\newblock {\em SIAM Journal on Scientific Computing}, 43(5):A3055--A3081, 2021.

\bibitem{lagari2020systematic}
Pola~Lydia Lagari, Lefteri~H Tsoukalas, Salar Safarkhani, and Isaac~E Lagaris.
\newblock Systematic construction of neural forms for solving partial differential equations inside rectangular domains, subject to initial, boundary and interface conditions.
\newblock {\em International Journal on Artificial Intelligence Tools}, 29(05):2050009, 2020.

\bibitem{dong2021method}
Suchuan Dong and Naxian Ni.
\newblock A method for representing periodic functions and enforcing exactly periodic boundary conditions with deep neural networks.
\newblock {\em Journal of Computational Physics}, 435:110242, 2021.

\bibitem{RN1920}
Kevin~Stanley McFall and James~Robert Mahan.
\newblock Artificial neural network method for solution of boundary value problems with exact satisfaction of arbitrary boundary conditions.
\newblock {\em IEEE Transactions on Neural Networks}, 20(8):1221--1233, 2009.

\bibitem{berg2018unified}
Jens Berg and Kaj Nystr{\"o}m.
\newblock A unified deep artificial neural network approach to partial differential equations in complex geometries.
\newblock {\em Neurocomputing}, 317:28--41, 2018.

\bibitem{RN1955}
Petr Karnakov, Sergey Litvinov, and Petros Koumoutsakos.
\newblock Solving inverse problems in physics by optimizing a discrete loss: Fast and accurate learning without neural networks.
\newblock {\em PNAS Nexus}, 2024.

\bibitem{zhang2021hierarchical}
Lei Zhang, Lin Cheng, Hengyang Li, Jiaying Gao, Cheng Yu, Reno Domel, Yang Yang, Shaoqiang Tang, and Wing~Kam Liu.
\newblock Hierarchical deep-learning neural networks: finite elements and beyond.
\newblock {\em Computational Mechanics}, 67:207--230, 2021.

\bibitem{lu2023convolution}
Ye~Lu, Hengyang Li, Lei Zhang, Chanwook Park, Satyajit Mojumder, Stefan Knapik, Zhongsheng Sang, Shaoqiang Tang, Daniel~W Apley, Gregory~J Wagner, et~al.
\newblock Convolution hierarchical deep-learning neural networks (c-hidenn): finite elements, isogeometric analysis, tensor decomposition, and beyond.
\newblock {\em Computational Mechanics}, 72(2):333--362, 2023.

\bibitem{zhang2022hidenn}
Lei Zhang, Ye~Lu, Shaoqiang Tang, and Wing~Kam Liu.
\newblock Hidenn-td: reduced-order hierarchical deep learning neural networks.
\newblock {\em Computer Methods in Applied Mechanics and Engineering}, 389:114414, 2022.

\bibitem{lu2021learning}
Lu~Lu, Pengzhan Jin, Guofei Pang, Zhongqiang Zhang, and George~Em Karniadakis.
\newblock Learning nonlinear operators via deeponet based on the universal approximation theorem of operators.
\newblock {\em Nature machine intelligence}, 3(3):218--229, 2021.

\bibitem{li2020fourier}
Zongyi Li, Nikola Kovachki, Kamyar Azizzadenesheli, Burigede Liu, Kaushik Bhattacharya, Andrew Stuart, and Anima Anandkumar.
\newblock Fourier neural operator for parametric partial differential equations.
\newblock {\em arXiv preprint arXiv:2010.08895}, 2020.

\bibitem{salcedo2014support}
Sancho Salcedo-Sanz, Jos{\'e}~Luis Rojo-{\'A}lvarez, Manel Mart{\'\i}nez-Ram{\'o}n, and Gustavo Camps-Valls.
\newblock Support vector machines in engineering: an overview.
\newblock {\em Wiley Interdisciplinary Reviews: Data Mining and Knowledge Discovery}, 4(3):234--267, 2014.

\bibitem{owhadi2019statistical}
Houman Owhadi, Clint Scovel, and Florian Sch{\"a}fer.
\newblock Statistical numerical approximation.
\newblock {\em Notices of the AMS}, 2019.

\bibitem{RN1873}
Jiahao Zhang, Shiqi Zhang, and Guang Lin.
\newblock Pagp: A physics-assisted gaussian process framework with active learning for forward and inverse problems of partial differential equations.
\newblock {\em arXiv preprint arXiv:2204.02583}, 2022.

\bibitem{RN1919}
Tomoharu Iwata and Zoubin Ghahramani.
\newblock Improving output uncertainty estimation and generalization in deep learning via neural network gaussian processes.
\newblock {\em arXiv preprint arXiv:1707.05922}, 2017.

\bibitem{RN1890}
Rui Meng and Xianjin Yang.
\newblock Sparse gaussian processes for solving nonlinear pdes.
\newblock {\em Journal of Computational Physics}, 490:112340, 2023.

\bibitem{RN1886}
Yifan Chen, Bamdad Hosseini, Houman Owhadi, and Andrew~M Stuart.
\newblock Solving and learning nonlinear pdes with gaussian processes.
\newblock {\em Journal of Computational Physics}, 447:110668, 2021.

\bibitem{RN1881}
Pau Batlle, Matthieu Darcy, Bamdad Hosseini, and Houman Owhadi.
\newblock Kernel methods are competitive for operator learning.
\newblock {\em Journal of Computational Physics}, 496, 2024.

\bibitem{wang2023discovery}
Kang Wang, Lei Zhang, and Shaoqiang Tang.
\newblock Discovery of pdes driven by data with sharp gradient or discontinuity.
\newblock {\em Computers \& Mathematics with Applications}, 140:33--43, 2023.

\bibitem{RN332}
Carl~Edward Rasmussen.
\newblock {\em Gaussian processes for machine learning}.
\newblock 2006.

\bibitem{RN1935}
Amin Yousefpour, Zahra~Zanjani Foumani, Mehdi Shishehbor, Carlos Mora, and Ramin Bostanabad.
\newblock Gp+: A python library for kernel-based learning via gaussian processes.
\newblock {\em arXiv preprint arXiv:2312.07694}, 2023.

\bibitem{RN1927}
Jacob Gardner, Geoff Pleiss, Kilian~Q Weinberger, David Bindel, and Andrew~G Wilson.
\newblock Gpytorch: Blackbox matrix-matrix gaussian process inference with gpu acceleration.
\newblock {\em Advances in neural information processing systems}, 31, 2018.

\bibitem{RN783}
R.~Bostanabad, T.~Kearney, S.~Y. Tao, D.~W. Apley, and W.~Chen.
\newblock Leveraging the nugget parameter for efficient gaussian process modeling.
\newblock {\em International Journal for Numerical Methods in Engineering}, 114(5):501--516, 2018.

\bibitem{RN940}
R.~Bostanabad, Y.~C. Chan, L.~W. Wang, P.~Zhu, and W.~Chen.
\newblock Globally approximate gaussian processes for big data with application to data-driven metamaterials design.
\newblock {\em Journal of Mechanical Design}, 141(11), 2019.

\bibitem{RN1559}
N.~Oune and R.~Bostanabad.
\newblock Latent map gaussian processes for mixed variable metamodeling.
\newblock {\em Computer Methods in Applied Mechanics and Engineering}, 387:114128, 2021.

\bibitem{RN434}
Matthew Plumlee and Daniel~W. Apley.
\newblock Lifted brownian kriging models.
\newblock {\em Technometrics}, 59(2):165--177, 2017.

\bibitem{RN1912}
Liang Ding, Simon Mak, and CF~Wu.
\newblock Bdrygp: a new gaussian process model for incorporating boundary information.
\newblock {\em arXiv preprint arXiv:1908.08868}, 2019.

\bibitem{RN1573}
Liwei Wang, Suraj Yerramilli, Akshay Iyer, Daniel Apley, Ping Zhu, and Wei Chen.
\newblock Scalable gaussian processes for data-driven design using big data with categorical factors.
\newblock {\em Journal of Mechanical Design}, 144(2), 2021.

\bibitem{RN1901}
Andrew~Gordon Wilson, Zhiting Hu, Ruslan Salakhutdinov, and Eric~P Xing.
\newblock Deep kernel learning.
\newblock In {\em Artificial intelligence and statistics}, pages 370--378. PMLR, 2016.

\bibitem{paszke2019pytorch}
Adam Paszke, Sam Gross, Francisco Massa, Adam Lerer, James Bradbury, Gregory Chanan, Trevor Killeen, Zeming Lin, Natalia Gimelshein, Luca Antiga, et~al.
\newblock Pytorch: An imperative style, high-performance deep learning library.
\newblock {\em Advances in neural information processing systems}, 32, 2019.

\bibitem{ohwada2009cole}
Taku Ohwada.
\newblock Cole-hopf transformation as numerical tool for the burgers equation.
\newblock {\em Appl. Comput. Math}, 8(1):107--113, 2009.

\bibitem{multiphysics1998introduction}
COMSOL Multiphysics.
\newblock Introduction to comsol multiphysics{\textregistered}.
\newblock {\em COMSOL Multiphysics, Burlington, MA, accessed Feb}, 9(2018):32, 1998.

\bibitem{arora2004introduction}
Jasbir~Singh Arora.
\newblock {\em Introduction to optimum design}.
\newblock Elsevier, 2004.

\bibitem{vaswani2017attention}
Ashish Vaswani, Noam Shazeer, Niki Parmar, Jakob Uszkoreit, Llion Jones, Aidan~N Gomez, {\L}ukasz Kaiser, and Illia Polosukhin.
\newblock Attention is all you need.
\newblock {\em Advances in neural information processing systems}, 30, 2017.

\bibitem{saha2021hierarchical}
Sourav Saha, Zhengtao Gan, Lin Cheng, Jiaying Gao, Orion~L Kafka, Xiaoyu Xie, Hengyang Li, Mahsa Tajdari, H~Alicia Kim, and Wing~Kam Liu.
\newblock Hierarchical deep learning neural network (hidenn): an artificial intelligence (ai) framework for computational science and engineering.
\newblock {\em Computer Methods in Applied Mechanics and Engineering}, 373:113452, 2021.

\bibitem{yousefpour2024simultaneous}
Amin Yousefpour, Shirin Hosseinmardi, Carlos Mora, and Ramin Bostanabad.
\newblock Simultaneous and meshfree topology optimization with physics-informed gaussian processes.
\newblock {\em arXiv preprint arXiv:2408.03490}, 2024.

\bibitem{batlle2024kernel}
Pau Batlle, Matthieu Darcy, Bamdad Hosseini, and Houman Owhadi.
\newblock Kernel methods are competitive for operator learning.
\newblock {\em Journal of Computational Physics}, 496:112549, 2024.

\bibitem{kingma2014adam}
Diederik~P Kingma and Jimmy Ba.
\newblock Adam: A method for stochastic optimization.
\newblock {\em arXiv preprint arXiv:1412.6980}, 2014.

\bibitem{liu1989limited}
Dong~C Liu and Jorge Nocedal.
\newblock On the limited memory bfgs method for large scale optimization.
\newblock {\em Mathematical programming}, 45(1-3):503--528, 1989.

\bibitem{sun2019survey}
Shiliang Sun, Zehui Cao, Han Zhu, and Jing Zhao.
\newblock A survey of optimization methods from a machine learning perspective.
\newblock {\em IEEE transactions on cybernetics}, 50(8):3668--3681, 2019.

\bibitem{scholkopf2002learning}
Bernhard Sch{\"o}lkopf and Alexander~J Smola.
\newblock {\em Learning with kernels: support vector machines, regularization, optimization, and beyond}.
\newblock MIT press, 2002.

\bibitem{o1986automatic}
Finbarr O'sullivan, Brian~S Yandell, and William~J Raynor~Jr.
\newblock Automatic smoothing of regression functions in generalized linear models.
\newblock {\em Journal of the American Statistical Association}, 81(393):96--103, 1986.

\bibitem{kimeldorf1971some}
George Kimeldorf and Grace Wahba.
\newblock Some results on tchebycheffian spline functions.
\newblock {\em Journal of mathematical analysis and applications}, 33(1):82--95, 1971.

\bibitem{szeliski1987regularization}
Richard Szeliski.
\newblock Regularization uses fractal priors.
\newblock In {\em Proceedings of the sixth National conference on Artificial intelligence-Volume 2}, pages 749--754, 1987.

\bibitem{liu2024discontinuity}
Li~Liu, Shengping Liu, Hui Xie, Fansheng Xiong, Tengchao Yu, Mengjuan Xiao, Lufeng Liu, and Heng Yong.
\newblock Discontinuity computing using physics-informed neural networks.
\newblock {\em Journal of Scientific Computing}, 98(1):22, 2024.

\bibitem{lax2005systems}
Peter Lax and Burton Wendroff.
\newblock Systems of conservation laws.
\newblock In {\em Selected Papers Volume I}, pages 263--283. Springer, 2005.

\end{thebibliography}
\end{document}